\documentclass[runningheads]{llncs}

\usepackage{times}
\usepackage{epsfig}
\usepackage{graphicx}
\usepackage{amsmath}
\usepackage{amssymb}
\usepackage{booktabs}
\usepackage{multirow}
\usepackage{caption}
\usepackage{subcaption}
\usepackage{wrapfig}
\usepackage[table,xcdraw]{xcolor}
\usepackage{comment}
\usepackage{color}
\usepackage{tikzpagenodes}
\usepackage[toc,page]{appendix}

\usepackage[pagebackref=true,breaklinks=true,letterpaper=true,colorlinks,bookmarks=false]{hyperref}

\newcommand{\datasets}{Pose-Audio-Transcript-Style (PATS) dataset }
\newcommand{\datasetss}{Pose-Audio-Transcript-Style (PATS)}

\newcommand{\datasetshorts}{PATS dataset }
\newcommand{\datasetshortss}{PATS}
\newcommand{\datasetshortsss}{PATS }

\newcommand{\Arrow}[1]{\tikz[baseline=-0.5ex]{\draw[#1, very thick] [->](0,0) -- (2ex,0); }}

\newcommand{\modelshort}{Mix-StAGE}
\newcommand{\modelshorts}{Mix-StAGE }

\newcommand{\set}[2]{\lbrace #1^1, #1^2, \ldots #1^#2\rbrace}
\newcommand{\setLow}[2]{\lbrace #1_1, #1_2, \ldots #1_#2\rbrace}

\newcommand{\makered}[1]{{\color{red}#1}}
\begin{document}
% \linenumbers
\pagestyle{headings}
\mainmatter
\def\ECCVSubNumber{2962}  % Insert your submission number here

%\title{Author Guidelines for ECCV Submission} % Replace with your title
%\title{Few-shot Learning for Individual Styles of Co-speech Gesture Animation}
%\title{Few Shot Learning of Style Transfer \\for Co-speech Gesture Animation}
\title{Style Transfer for Co-Speech Gesture Animation:\\ A Multi-Speaker Conditional-Mixture Approach}
\begin{comment}
\titlerunning{ECCV-20 submission ID \ECCVSubNumber} 
\authorrunning{ECCV-20 submission ID \ECCVSubNumber} 
\author{Anonymous ECCV submission}
\institute{Paper ID \ECCVSubNumber}
\end{comment}
%******************

%\begin{comment}
\titlerunning{Style Transfer for Co-Speech Gesture Animation}
\author{Chaitanya Ahuja\inst{1}\orcidID{0000-0003-4396-2050} \and
Dong Won Lee\inst{1} \and
Yukiko I. Nakano\inst{2} \and
Louis-Philippe Morency\inst{1}\orcidID{0000-0001-6376-7696}}
%% Name Parsing for authors
%% Dong Won, Lee 
%% Louis-Philippe Morency
%
\authorrunning{C. Ahuja et al.}
\institute{Carnegie Mellon University, Pittsburgh, PA, USA
\email{\{cahuja,dongwonl,morency\}@cs.cmu.edu}\\ \and
	Seikei University, Musashino, Tokyo, Japan \\
\email{y.nakano@st.seikei.ac.jp}}
%\end{comment}
%******************
\maketitle

%%%%%%%%% TITLE

% \and

%\maketitle
%\thispagestyle{empty}
% \twocolumn[{%
% \renewcommand\twocolumn[1][]{#1}%
% \maketitle
% \vspace{-1.5cm}
% \begin{center}
%     \centering
%     \includegraphics[width=0.9\textwidth]{figs/overview.png}
%     \captionof{figure}{Overview of cross-modal translation task which uses aligned language and speech to generate plausible pose sequences. The generated poses belong to different modes in the underlying distribution.}\label{fig:overview}
% \end{center}%
% }]

%%%%%%%%% ABSTRACT
\begin{abstract}
\textit{How can we teach robots or virtual assistants to gesture naturally? Can we go further and adapt the gesturing style to follow a specific speaker?} Gestures that are naturally timed with corresponding speech during human communication are called co-speech gestures. A key challenge, called gesture style transfer, is to learn a model that generates these gestures for a speaking agent `A' in the gesturing style of a target speaker `B'. A secondary goal is to simultaneously learn to generate co-speech gestures for multiple speakers while remembering what is unique about each speaker. We call this challenge style preservation. 
%in different styles while speaking and communicating with humans?} These co-speech gestures need to be properly timed with the corresponding audio and have the correct style based on the context. 
%Some recent work has proposed neural architectures that can map speech to gestures, but these approaches learn speaker-specific models. 
%A key challenge is to model co-speech gestures of multiple speakers while preserving the style of each speaker. Another challenge is to alter the style of the generated gestures (i.e. transfer gesture style from one speaker to the other). 
In this paper, we propose a new model, named \modelshort, which trains a single model for multiple speakers while learning unique style embeddings for each speaker's gestures in an end-to-end manner. A novelty of \modelshorts is to learn a mixture of generative models which allows for conditioning on the unique gesture style of each speaker.
%while the audio drives the co-speech gesture generation. 
As \modelshorts disentangles style and content of gestures, gesturing styles for the same input speech can be altered by simply switching the style embeddings. \modelshorts also allows for style preservation when learning simultaneously from multiple speakers. We also introduce a new dataset, \datasetss, designed to study gesture generation and style transfer. 
%It consists of 25 speakers (15 new speakers and 10 speakers from Ginosar et al. 2019) for a total of 250+ hours of gestures and aligned audio signals. 
Our proposed \modelshorts model significantly outperforms the previous state-of-the-art approach for gesture generation and provides a path towards performing gesture style transfer across multiple speakers. Link to code, data and videos: \url{http://chahuja.com/mix-stage}.
%\dots
%\vspace{-0.2cm}
%to include a mixture model which is guided by multiple mode priors from the speaker’s pose probability distribution. This allows \modelshorts to generate diverse, expressive and natural looking gestures. 

% Most speakers will have multiple prototypical gestures, which represent distinct modes in their gesture distribution
\end{abstract}

%%%%%%%%% BODY TEXT
\section{Introduction}
\label{sec:intro}
\begin{figure}
    \centering
    \includegraphics[width=\textwidth]{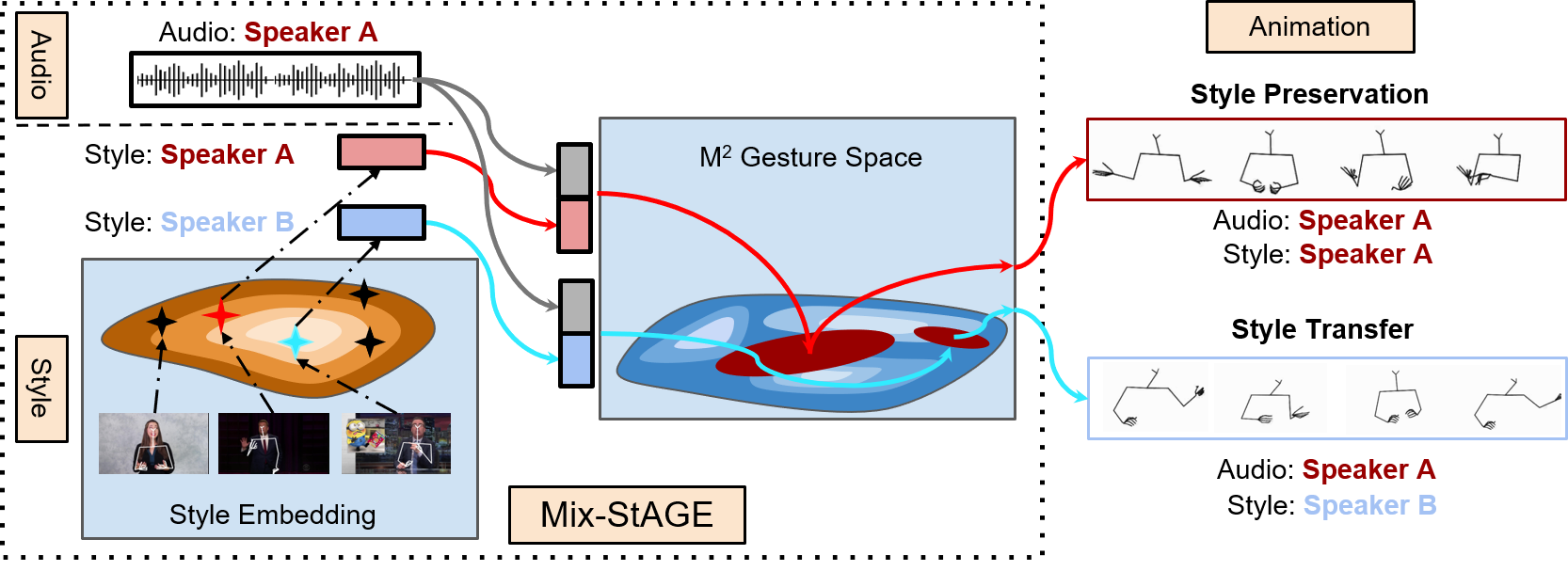}
    \caption{Overview of co-speech gesture generation and gesture style transfer/preservation task. The models learns a style embedding for each speaker, which can be be mapped to a gesture space with either the same speaker's audio to generate style preserved gestures or a different speaker's audio to generate style transferred gestures.}
    \label{fig:overview}
    %\vspace{-3em}
\end{figure}
Nonverbal behaviours such as body posture, hand gestures and head nods play a crucial role in human communication \cite{wagner2014gesture,obermeier2015speaker}. Pointing at different objects, moving hands up-down in emphasis, and describing the outline of a shape are some of the many gestures that co-occur with the verbal and vocal content of communication. These are known as co-speech gestures~\cite{McNeill1992,Kendon1980}. When creating new robots or embodied virtual assistants designed to communicate with humans, it is important to generate \textit{naturalistic} looking gestures that are meaningful with the speech \cite{bailenson2006effect}. Some recent works have proposed speaker-specific gesture generation models \cite{ginosar2019learning,Chiu2015,cassell2004beat,ferstl2019multi} that are both trained and tested on the same speaker. The intuition behind this prior work is that co-speech gestures are idiosyncratic \cite{xu2009symbolic,McNeill1992}. There is an unmet need to learn generative models that are able to learn to generate gestures simultaneously from multiple speakers (\Arrow{red} in Figure \ref{fig:overview}) while at the same time remembering what is unique for each speaker's gesture style. These models should not simply remember the ``average'' speaker. A bigger technical challenge is to be able to transfer gesturing style of speaker `B' to speaker `A' (\Arrow{blue} in Figure \ref{fig:overview}).
%We argue that even though gestures of different speakers are different, they are not completely independent (a subset of gestures of one speaker might also be made by another speaker). The model trained on multiple speakers would benefit from these common gestures across different speakers. 
%A key challenge to modeling multiple speakers jointly is preserving the style of each speaker, while modeling co-speech gestures at the same time. 

The gesturing style can defined along two dimensions which is a result of (a) the speaker’s idiosyncrasy (or speaker-level style), and (b) due to some more general attributes such as standing versus sitting, or the body orientation such as left versus right (or attribute-level style). For both gesture style types, the generation model needs to be able to learn the diversity and expressivity \cite{pelachaud2009studies,Bergmann2009} present in the gesture space, within and amongst speakers. The gesture distribution is likely to have multiple modes, some of them shared among speakers and some distinct to a speaker’s prototypical gestures.

In this paper, we introduce the \underline{Mix}ture-Model guided \underline{St}yle and \underline{A}udio for \underline{Ge}sture Generation (or \modelshort) approach which trains a single model for multiple speakers while learning unique style embeddings for each speaker's gestures in an end-to-end manner (see Figure \ref{fig:overview}). We use this model to perform two tasks for gesture generation conditioned on the input audio signal, (1) \textbf{style preservation} which ensures that while learning from multiple speakers we are still able to preserve unique gesturing styles of each speaker, and (2) \textbf{style transfer} where generated gestures are from a new style that was not the same as the source of the speech. A novelty of \modelshorts is to learn a mixture of generative models which allows for conditioning on the unique gesture style of each speaker. Our experiments study the impact of multiple speakers on both style transfer and preservation. Our study focuses on the non-verbal components of speech asking the research question if we can predict gestures without explicitly modeling verbal signals. We also introduce a new dataset, \datasetss, designed to study gesture generation and style transfer. %It consists of 25 speakers (15 new speakers and 10 speakers from Ginosar \cite{ginosar2019learning}) for a total of 250+ hours of gestures and aligned audio signals.
%\begin{wrapfigure}{r}{0.5\textwidth}
\begin{figure}[ht]
\begin{center}
    \includegraphics[width=\linewidth]{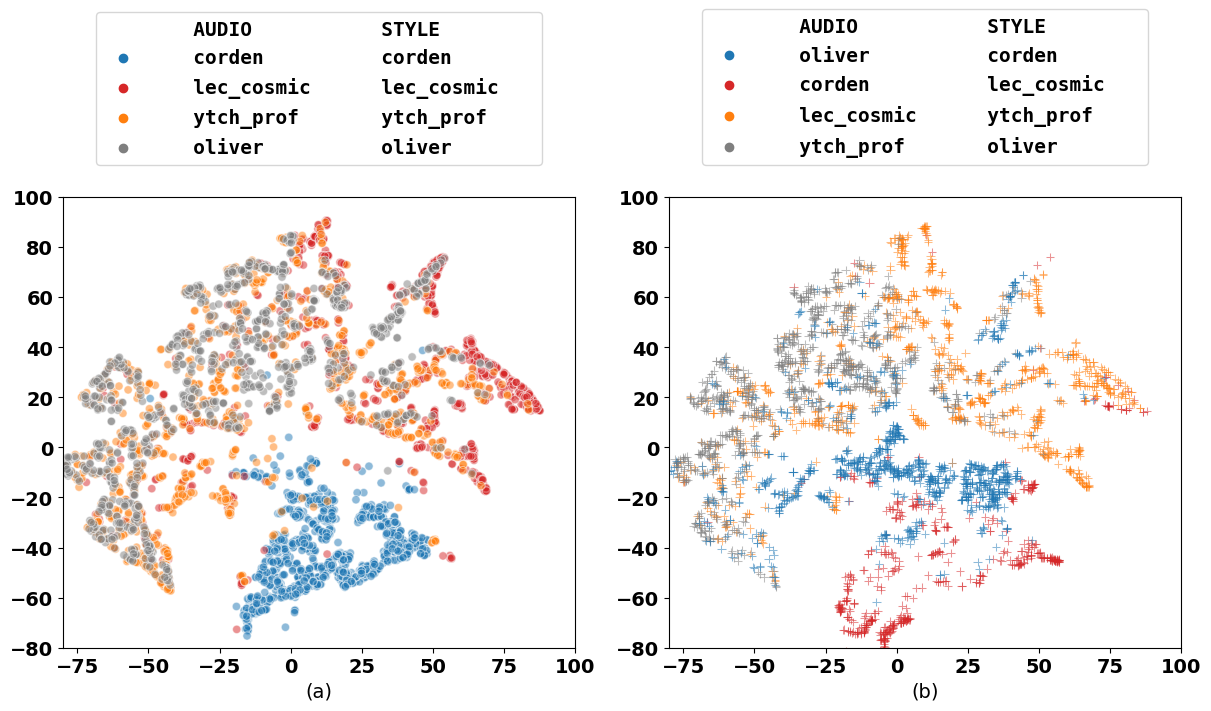}
\end{center}
\caption{t-SNE\cite{maaten2008visualizing} representation of the Multi-mode Multimodal Gesture Space (Section \ref{ssec:cluster}). Each color represents a style, which is fixed for both plots. The plot on the left visualizes the gesture space generated from the audio content and style of the same speaker. The plot on the right shows the generated gesture space where the audio content and style are not from the same speaker. It can be observed that a similar gesture space is occupied by each speaker's style even when the audio content is not of their own.}
\vspace{-1.5em}
\label{fig:tsne}
\end{figure}

\section{Related Work}
\label{sec:related}

\noindent \textbf{Speech driven Gesture Generation}:
For prosody-driven head motion generation \cite{Sargin2008} and body motion generation \cite{Levine2009,Levine2010}, Hidden Markov Models were used to predict a sequence of frames. Chiu \& Marsella \cite{Chiu2014} proposed a two-step process: predicting gesture labels from speech signal using conditional random fields (CRFs) and converting the label to gesture motion using Gaussian process latent variable models (GPLVMs). More recently, an LSTM network was applied to MFCC features extracted from speech to predict a sequence of frames for gestures \cite{Hasegawa2018} and body motions \cite{Shlizerman2018,ahuja2019react}. Generative adversarial networks (GAN) were used to generate head motions \cite{Sadoughi2018} and body motions\cite{ferstl2019multi}. Gestures driven by an audio signal\cite{ginosar2019learning} is the closest approach to our task of style preservation but it uses models trained on single speakers unlike our multi-speaker models. %Generalizability of these speaker-specific models are limited and requires large amounts of training data to learn a new speakers' models.
%It is shown that gesture generation using adversarial methods can benefit from the start-time of the gestures\cite{ferstl2019multi}, but such methods require large-scale annotations which are sometimes not feasible.
\subsubsection{Disentanglement and Transfer of Style}:
Style extraction and transfer have been studied in context of image artistic style \cite{gatys2015neural,johnson2016perceptual}, factorizing foreground and background in videos\cite{denton2017unsupervised,villegas2017decomposing}, disentanglement in speech \cite{wang2018style,bian2019multi,gurunath2019disentangling}. These approaches were extended to translation between properties of style such as map edges and real photos using paired samples \cite{isola2017image}. Paired data limits the variety of attributes of source and target, which encouraged unsupervised domain translation for images \cite{zhu2017unpaired,zhu2017toward} and videos\cite{bansal2018recycle}. Style was disentangled from content using a shared latent space\cite{liu2017unsupervised}, a cycle consistency loss \cite{zhu2017unpaired} and contrastive learning \cite{nagrani2020disentangled}. Cycle consistency losses were shown to limit diversity in the generated outputs as opposed to a weak consistency loss \cite{huang2018multimodal} and shared content space \cite{lee2019drit++}. Cycle consistency in cross-domain translation assumes reversibility (i.e. domain A can be translated to domain B and vice-versa). These assumptions are violated in cross-modal translation \cite{ma2018neural} and style control \cite{wang2018style} tasks where information in modality B (e.g. pose) is a subset of that in modality B (e.g. audio).
Style transfer for pose has been studied in context of generating dance moves based on the content of the audio \cite{lee2019dancing} or walking styles \cite{smith2019efficient}. Generated dance moves are conditioned on both the style and content of the audio (i.e. kind of music like ballet or hip-hop), unlike co-speech gesture generation which requires only the content and not the style of the audio (i.e. speaker specific style like identity or fundamental frequency). Co-speech gesture styles have been studied in context of speaker personalities \cite{neff2008gesture}, but requires a long annotation process to create a profile for each speaker. To our knowledge, this is the first fully data-driven approach that learns gesture style transfer for multiple speakers in a co-speech gesture generation setting.

%\input{z84_tsne_pair.tex}
%\input{z82_table_main}
%\input{z83_table_evil}

%\input{z88_multimode.tex}
%\input{z88_multimode.tex}
%\input{z90_clusters.tex}
% \section{\datasets}
% \label{sec:dataset}
% \input{4_dataset}

\section{Stylized Co-Speech Gesture Animation}
\label{sec:problem}
We define the problem of stylized co-speech gesture animation with two main goals, (1) generation of an animation which represents the gestures that would co-occur with the the spoken segment and (2) modification of the style of these gestures. Figure \ref{fig:overview} shows the first goal (\Arrow{red}) exemplified with the style preservation scenario, while the second goal (\Arrow{blue}) exemplifies with the style transfer scenario.

Formally, given a sequence of T audio frames $\textbf{X}_a \sim F_a$ and $i^{th}$ speaker's style $S(i)$, the goal is to predict a sequence of T frames of 2-D poses $\textbf{Y}_p \sim F_p$. Here $F_a$ and $F_p$ are the marginal distributions of the content of input audio and style of output pose sequences. To control pose generation by both style and audio, we learn a joint distribution over pose, audio and style $F_{p, a, s}$ which can be broken down into 3 parts

\begin{equation}
    F_{p, a, s} = F_{p|\Phi} F_{\Phi|a, s} \cdot F_{s}\cdot F_{a}
\end{equation}
where $F_{\Phi|a, s}$ is the distribution of the gesture space $\Phi$ conditioned on the audio and style of pose (Figure \ref{fig:overview}). We discuss the modelling of $F_{p|\Phi} F_{\Phi|a, s}$, $F_{a}$, and  $F_{s}$ in Section  \ref{ssec:cluster}, \ref{ssec:recon} and \ref{ssec:style} respectively.

\begin{figure}[h]
\centering
\includegraphics[width=\linewidth]{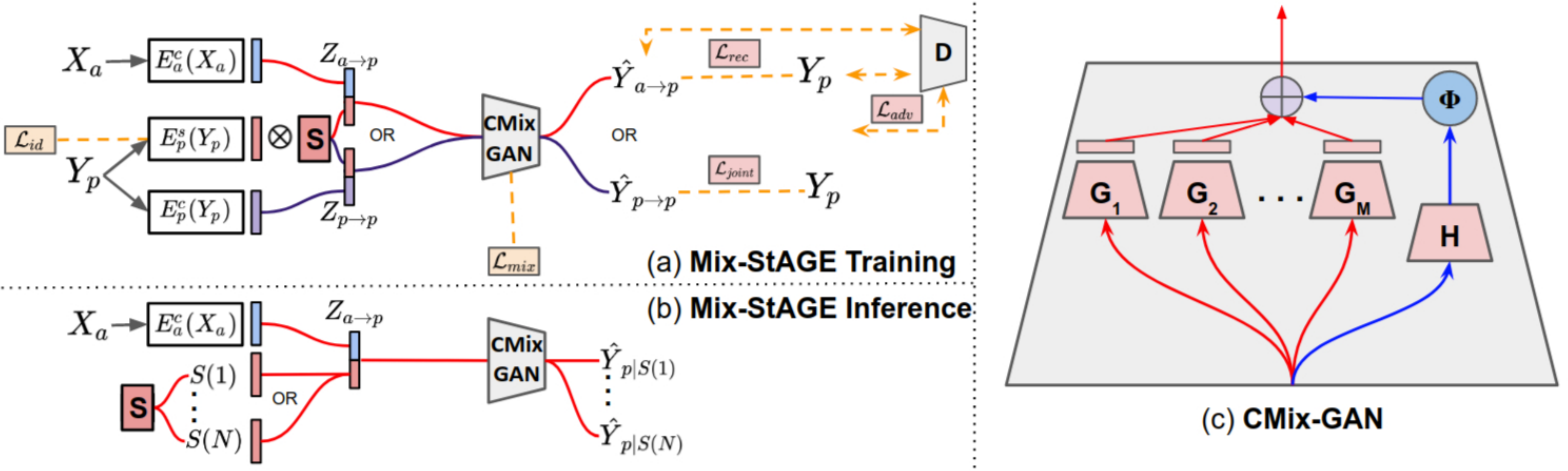}
\caption{(a) Overview of the proposed model \modelshorts in training mode, where audio $X_a$ and pose $Y_p$ are fed as inputs to learn a style embedding and concurrently generate a gesture animation. $\textbf{S}$ represents the style matrix, which is multiplied with a separately encoded pose. $\bigotimes$ represents $\text{argmax}$ for style ID followed by matrix multiplication. Discriminator D is used for adversarial training. All the loss functions are represented with dashed lines. (b) \modelshorts in inference mode, where any speaker's style embedding can be used on an input audio $X_a$ to generate gesture style-transferred or style-preserved animations (c) CMix-GAN generator: a visual representation of the conditional Mix-GAN model, where the $\bigoplus$ represents a weighted sum of the model priors $\Phi$ with the generated outputs by the sub-generators. }

% phi for training instance where only style and pose is used?
\label{fig:model}
\end{figure}
\section{Mix-StAGE: \underline{Mix}ture-Model guided \underline{St}yle and \underline{A}udio for \underline{Ge}sture Generation}
\label{sec:model}
% \comm{\cite{liu2016coupled} sharing weights for a part of the encoder helps with decoupling 
% \comm{a cross-modal translation is not reversible in general, as pose sequences do not have enough information to generate back the corresponding audio. Hence a bi-directional reconstruction loss for latent distribution matching can't be employed. We use a joint loss which achieves the same goal of latent distribution matching for a cross-modal translation task with unbalanced modalities.}
% \comm{We argue that speakers' audio signals have implicit information about the identity of the speaker}
% style and content, that is why part of our encoders have shared weights.} 
 
Figure \ref{fig:model} shows an overview of our \modelshorts model, including the training inference pathways. A first component of our \modelshorts model is the audio encoder $E^c_a$, which takes as input the spoken audio $X_a$. During training, we also have the pose sequence of the speaker $Y_p$. This pose sequence is decomposed into content and style, with two specialized encoders $E^c_p$ and $E^s_p$. During training, the style for the pose sequence can either be concatenated with the audio or the pose content.

% \begin{align}
%   \textbf{Z}_{a\rightarrow p} = \left[ E^c_a(\textbf{X}_a), E^s_p(\textbf{Y}_p) \right]\\
%   \textbf{Z}_{p\rightarrow p} = \left[ E^c_p(\textbf{Y}_p), E^s_p(\textbf{Y}_p) \right]
% \end{align}
The pose sequences for multiple speakers are represented as a distribution with multiple modes \cite{hao2018mixgan}. To decode from this multi-mode multimodal gesture space, we use a common generator $G$ with multiple sub-generators (or CMix-GAN) conditioned on input audio and style to decode both these embeddings to output pose $\textbf{Y}_p$. 

Our loss function comprises of a mode-regularization loss (Section \ref{ssec:cluster}) to ensure that audio and style embedding can sample from the appropriate marginal distribution of poses, a joint loss (Section \ref{ssec:recon}) to ensure latent distribution matching for content in a cross-modal translation task, a style consistency loss (Section \ref{ssec:style}) to ensure that the correct style is being generated and an adversarial loss (Section \ref{ssec:adv}) that matches the generated pose distribution to the target pose distribution.

\subsection{M$^2$GS: Multi-mode Multimodal Gesture Space }
\label{ssec:cluster}
Humans perform different styles of gestures, where each style consists of  different kinds of gestures (i.e beat, metaphorical, emblematic, iconic and so on)\cite{McNeill1992}. Learning pose generators for multiple speakers, each with their own style of gestures, presents a distribution with multiple modes. These gestures have a tendency of switching from one mode to the other over time, which depends on style embeddings and content of the audio. 

To prevent mode collapse \cite{arjovsky2017wasserstein} we propose the use of mixture-model guided sub-generators \cite{hao2018mixgan,arora2017generalization,hoang2018mgan}, each learning a different mode of $M^2$ gesture space $F_{\Phi|a,s}$.
\begin{equation}
    \mathbf{\hat{Y}}_p = \sum_{m=1}^M \mathbf{\phi}_m G_m(\mathbf{Z}) = G(\textbf{Z})\label{eq:mm_func}
\end{equation}
where $\textbf{Z} \in \lbrace \textbf{Z}_{a\rightarrow p}, \textbf{Z}_{p\rightarrow p}\rbrace$ are cross-modal and self-modal latent spaces respectively. They are defined as $\textbf{Z}_{a\rightarrow p} = \left[ E^c_a(\textbf{X}_a), E^s_p(\textbf{Y}_p)\bigotimes \textbf{S} \right]$ and $\textbf{Z}_{p\rightarrow p} = \left[ E^c_p(\textbf{Y}_p), E^s_p(\textbf{Y}_p)\bigotimes \textbf{S} \right]$ where $\textbf{S}$ is the style embedding matrix (See Section \ref{ssec:style}) and $\bigotimes$ is $\text{argmax}$ for style ID followed by matrix multiplication. Pose sequence $\hat{Y}_p \sim F_{p|\Phi}F_{\Phi|a,s}$ represents the pose probability distribution conditioned on audio and style. $G_m \sim F_{p|a, s}^m \hspace{0.1cm}\forall m \in \left[1, 2, \ldots M\right]$ are sub-generator functions with corresponding mixture-model priors $\Phi = \setLow{\mathbf{\phi}}{M}$. These mixture model priors represent the $M^2$ gesture space and are estimated at inference time conditioned on the input audio and style.

\noindent\textbf{Estimating Mixture Model Priors: }
During training, we partition poses $Y_p$ into $M$ clusters using an unsupervised approach, Lloyd's algorithm \cite{lloyd1982least}. While other unsupervised clustering methods \cite{reynolds2009gaussian} can also be used at this stage, we choose Lloyd's algorithm for its simplicity and speed. Each of these clusters represent samples from probability distributions $\set{F_{p|a, s}}{M}$. If a sample belongs to the $m^{th}$ cluster, $\phi_m = 1$, otherwise $\phi_m =0$, making $\Phi$ a sequence of one-hot vectors. While training the generator $G$ with loss function $\mathcal{L}_{\mbox{rec}}$, if a sample belongs to the distribution $F_{p|a, s}^m$, only parameters of sub-generator $G_m$ are updated. Hence, each sub-generator learns different components of the true distribution, which are combined using Equation \ref{eq:mm_func} to give the generated pose.

At inference time, we do not have the true values of mixture-model priors $\Phi$. As mixture model priors modulate based on the style of the speaker and audio content at any given moment, we jointly learn a classification network $H \sim F_{\Phi|a,s}$ to estimate values of $\Phi$ in form of a mode regularization loss function

%If $\Phi$ is learnt while training without any constraints, the generator does not have any incentive to learn multiple modes of the underlying probability distribution. Hence, this mixture model will become equivalent to having only one component which will again lead to mode collapse. To overcome this challenge, we use multiple modes in the distribution of true data which are incorporated as constraints while training.

\begin{equation}
    \mathcal{L}_{mix} = \mathbb{E}_{\Phi, \mathbf{Z}}\mbox{CCE}(\Phi, H(\mathbf{Z})) \label{eq:aux}
\end{equation}
where CCE is categorical cross-entropy. 
\subsection{Joint Space of Style, Audio and Pose}
\label{ssec:recon}
%These encoders learn to map to a cross-modal latent space $\textbf{Z}_{a\rightarrow p} = \left[ E^c_a(\textbf{X}_a), E^s_p(\textbf{Y}_p) \right]$. Cross-modal translation is not reversible for this task (i.e. audio signal cannot be generated with pose input). Hence a bi-directional reconstruction loss \cite{lee2019drit++} for latent distribution matching canott be employed. We use a self-modal latent space $\textbf{Z}_{p\rightarrow p} = \left[ E^c_p(\textbf{Y}_p), E^s_p(\textbf{Y}_p) \right]$ with a joint loss to resolve this dilemma. $E^c_p$ is the pose content encoder.
A set of marginal distributions $F_a$ and $F_s$ are learnt by our content encoders $E_a^c$ and $E_p^c$, which together define the joint distribution of the generated poses: $F_{p,a,s}$. Since both cross-modal $\textbf{Z}_{a\rightarrow p}$ and self-modal $\textbf{Z}_{p\rightarrow p}$ latent spaces are designed to represent the same underlying content distribution, they should be consistent with each other. Using the same generator $G$ for decoding both of these embeddings\cite{liu2016coupled} yields content invariant generator. We enforce a reconstruction and joint loss \cite{ahuja2019language2pose} which encourages a reduction in distance between $\textbf{Z}_{a\rightarrow p}$ and $\textbf{Z}_{p\rightarrow p}$. As cross-modal translation is not reversible for this task (i.e. audio signal cannot be generated with pose input), a bi-directional reconstruction loss \cite{lee2019drit++} for latent distribution matching cannot be directly used. This joint loss achieves the same goal of latent distribution matching in a uni-modal translation task \cite{huang2018multimodal,rosca2017variational,royer2020xgan} but for a cross-modal translation task. 

\begin{align}
    \mathcal{L}_{joint} &= \mathbb{E}_{\mathbf{Y}_p}\| \mathbf{Y}_p - G(\textbf{Z}_{p\rightarrow p})\|_1 \label{eq:joint} \\ 
    \mathcal{L}_{rec} &= \mathbb{E}_{\mathbf{Y}_p, \textbf{X}_a}\| \mathbf{Y}_p - G(\textbf{Z}_{a\rightarrow p})\|_1 \label{eq:reg}
\end{align}

\subsection{Style Embedding}
\label{ssec:style}
We represent style as a collection of embeddings $S(i) \in \textbf{S} \sim F_s$, where $S(i)$ is the style of the $i^{th}$ speaker in the style matrix $\textbf{S}$. Style space and embeddings are conceptually similar to the GST (Global Style Token) layer \cite{wang2018style} which decomposes the audio embedding space into a set of basis vectors or style tokens, but only one out of the two modalities in the stylized audio generation task \cite{wang2018style,ma2018neural} have both style and content. In our case, both audio and pose have style and content. To ensure that generator $G$ is attending only to style of pose while ignoring style of the audio, a style consistency loss is enforced on input $\textbf{Y}_p$ and generated $\hat{\textbf{Y}}_p$.

% One of the goals of this model is to control and modify the style of the generated poses while keeping its semantic meaning aligned with the content in the audio signal. Given that each speaker has their own unique style, a style embedding is learnt for each speaker solely based on their pose sequences. The collection of these style embeddings represent the style space $\textbf{S} \sim F_s$ where each row is an individualized style embedding. 

\begin{equation}
    \mathcal{L}_{id} = \mathbb{E}_{Y \in \lbrace\textbf{Y}_p, \hat{\textbf{Y}}_p\rbrace}\mbox{CCE}\left(\mbox{Softmax}\left(E^s_p(Y)\right), \textbf{ID}\right) \label{eq:style}
\end{equation}
where $\textbf{ID}$ is a one-hot vector denoting the speaker level style.

\subsection{Total Loss with Adversarial Training}
\label{ssec:adv}
%Solely, minimizing L1 distance between generated poses and ground-truth poses results in an overly smooth motion close to the mean of the probability distribution of poses ($F_p$) \cite{ginosar2019learning}. Using adversarial training \cite{goodfellow2014generative} with a discriminator on top of the decoder, cross-modal translation models can generate a better distribution of poses \cite{ginosar2019learning}.
To alleviate the challenge of overly smooth generation caused by L1 reconstruction and joint losses in Equation \ref{eq:joint},\ref{eq:reg}, we use the generated pose sequence $\mathbf{\hat{Y}}^p$ as a signal for the adversarial discriminator $D$ \cite{ginosar2019learning}. The discriminator tries to classify the true pose $\mathbf{Y}^p$ from the generated pose $\mathbf{\hat{Y}}^p$, while the generator learns to fool the discriminator by generating realistic poses. This adversarial loss\cite{goodfellow2014generative} is written as:
\begin{equation}
    \mathcal{L}_{adv} = \mathbb{E}_{\mathbf{Y}_p}\log D\left(\mathbf{Y}_p) + \mathbb{E}_{\textbf{X}_a, \textbf{Y}_p}\log\left(1- D(G\left(\left[E^c_a(\textbf{X}_a), E^s_p(\textbf{X}_p)\right]\right)\right)\right) \label{eq:adv}
\end{equation}

The model is jointly trained to optimize the overall loss function:
\begin{equation}
    \max_{D} \min_{E^c_a, E^c_p, E^s_p, G} \mathcal{L}_{mix} + \mathcal{L}_{joint} + \mathcal{L}_{rec} + \lambda_{id}\mathcal{L}_{id} + \mathcal{L}_{adv}  
\end{equation}
where $\lambda_{id}$ controls the weight of the style consistency loss term.
%in Equations \ref{eq:aux},\ref{eq:joint},\ref{eq:reg}, \ref{eq:style}, and \ref{eq:adv}

\subsection{Network Architectures}
Our proposed approach can work with any temporal network, giving it the flexibility of incorporating domain dependent or pre-trained temporal models. 
%Temporal Convolution Networks (TCNs) have been shown to perform well in speech-conditioned pose generation tasks \cite{ahuja2019react}. It has been shown that adding residual connections to TCNs in a one dimensional U-Net architecture \cite{ronneberger2015u} can help model global and local dynamics of input speech signals \cite{ginosar2019learning}. 

In our experiments we use a Temporal Convolution Network (TCN) module for both content and style encoders. The style space is a matrix $\mathbf{S}\in \mathbb{R}^{N \times D}$ where $N$ is the number of speakers and $D$ is the length of the style embeddings. The generator $G(.)$ consists of a 1D version of U-Net \cite{ronneberger2015u,ginosar2019learning} followed by $M$ TCNs as sub-generator functions. The discriminator is also a TCN module with lower capacity than the generators. A more detailed architecture can be found in the supplementary.%\footnote{Refer to supplementary for a detailed architecture.  [Link to code and dataset]}.
% \usepackage{booktabs}
% \usepackage{multirow}
\begin{table}[]
\centering
\begin{tabular}{@{}c|l|cc|cc|cc|cc|cc@{}}
\toprule
                                                                                      & \multicolumn{1}{c|}{}                                   & \multicolumn{4}{c|}{\textbf{Single-Speaker Models}}                                                                                & \multicolumn{6}{c|}{\textbf{Multi-Speaker Models}}                                                                                                                                                                 \\ \cmidrule(l){3-12} 
                                                                                      & \multicolumn{1}{c|}{}                                   & \multicolumn{2}{c|}{\textbf{S2G}\cite{ginosar2019learning}}                           & \multicolumn{2}{c|}{\textbf{CMix-GAN}}                               & \multicolumn{2}{c|}{\textbf{MUNIT}\cite{huang2018multimodal}}                         & \multicolumn{2}{c|}{\textbf{StAGE}}                                  & \multicolumn{2}{c|}{\textbf{Mix-StAGE}}                                       \\ \cmidrule(l){3-12} 
\multirow{-3}{*}{\textbf{\begin{tabular}[c]{@{}c@{}}No. of \\ Speakers\end{tabular}}} & \multicolumn{1}{c|}{\multirow{-3}{*}{\textbf{Speaker}}} & \textbf{PCK}                 & \textbf{F1}                  & \textbf{PCK}                 & \textbf{F1}                           & \textbf{PCK}                 & \textbf{F1}                  & \textbf{PCK}                 & \textbf{F1}                           & \textbf{PCK}                          & \textbf{F1}                           \\ \midrule
                                                                                      & \cellcolor[HTML]{EFEFEF}\textbf{Mean}                   & \cellcolor[HTML]{EFEFEF}0.25 & \cellcolor[HTML]{EFEFEF}0.08 & \cellcolor[HTML]{EFEFEF}0.26 & \cellcolor[HTML]{EFEFEF}\textbf{0.27} & \cellcolor[HTML]{EFEFEF}0.24 & \cellcolor[HTML]{EFEFEF}0.06 & \cellcolor[HTML]{EFEFEF}\textbf{0.36} & \cellcolor[HTML]{EFEFEF}0.21          & \cellcolor[HTML]{EFEFEF}0.34 & \cellcolor[HTML]{EFEFEF}0.22          \\ \cmidrule(lr){2-2}
                                                                                      & Corden                                                  & 0.30                         & 0.05                         & 0.32                         & 0.21                                  & 0.25                         & 0.06                         & 0.36                         & 0.21                                  & 0.34                                  & 0.24                                  \\
\multirow{-3}{*}{\textbf{2}}                                                          & lec\_cosmic                                             & 0.19                         & 0.12                         & 0.19                         & 0.33                                  & 0.15                         & 0.19                         & 0.20                         & 0.48                                  & 0.24                                  & 0.49                                  \\ \midrule
                                                                                      & \cellcolor[HTML]{EFEFEF}\textbf{Mean}                   & \cellcolor[HTML]{EFEFEF}0.37 & \cellcolor[HTML]{EFEFEF}0.18 & \cellcolor[HTML]{EFEFEF}0.37 & \cellcolor[HTML]{EFEFEF}0.27          & \cellcolor[HTML]{EFEFEF}0.22 & \cellcolor[HTML]{EFEFEF}0.05 & \cellcolor[HTML]{EFEFEF}0.38 & \cellcolor[HTML]{EFEFEF}\textbf{0.34} & \cellcolor[HTML]{EFEFEF}\textbf{0.39} & \cellcolor[HTML]{EFEFEF}\textbf{0.35} \\ \cmidrule(lr){2-2}
                                                                                      & Corden                                                  & 0.30                         & 0.05                         & 0.32                         & 0.21                                  & 0.24                         & 0.07                         & 0.35                         & 0.27                                  & 0.35                                  & 0.30                                  \\
\multirow{-3}{*}{\textbf{4}}                                                          & lec\_cosmic                                             & 0.19                         & 0.12                         & 0.19                         & 0.33                                  & 0.19                         & 0.16                         & 0.18                         & 0.23                                  & 0.20                                  & 0.19                                  \\ \midrule
                                                                                      & \cellcolor[HTML]{EFEFEF}\textbf{Mean}                   & \cellcolor[HTML]{EFEFEF}0.36 & \cellcolor[HTML]{EFEFEF}0.14 & \cellcolor[HTML]{EFEFEF}0.37 & \cellcolor[HTML]{EFEFEF}0.26          & \cellcolor[HTML]{EFEFEF}0.31 & \cellcolor[HTML]{EFEFEF}0.21 & \cellcolor[HTML]{EFEFEF}0.38 & \cellcolor[HTML]{EFEFEF}\textbf{0.32} & \cellcolor[HTML]{EFEFEF}\textbf{0.40} & \cellcolor[HTML]{EFEFEF}\textbf{0.33} \\ \cmidrule(lr){2-2}
                                                                                      & Corden                                                  & 0.30                         & 0.05                         & 0.32                         & 0.21                                  & 0.23                         & 0.03                         & 0.32                         & 0.28                                  & 0.36                                  & 0.27                                  \\
\multirow{-3}{*}{\textbf{8}}                                                          & lec\_cosmic                                             & 0.19                         & 0.12                         & 0.19                         & 0.33                                  & 0.13                         & 0.09                         & 0.23                         & 0.34                                  & 0.24                                  & 0.32                                  \\ \bottomrule
\end{tabular}
\caption{\textbf{Style Preservation}: Objective metrics for pose generation of single-speaker and multi-speaker models as indicated in the columns. Each row refers to the number of speakers the model was trained, with the average performance indicated at the top. The scores for common individual speakers are also indicated below alongside. For detailed results on other speakers please refer to the supplementary. Bold numbers indicate $p<0.1$ in a bootstrapped two sided t-test.}
\label{tab:main}
%\vspace{-3em}
\end{table}

\section{Experiments}
\label{sec:exp}
% Generating plausible pose sequences from speech and language signals can be broken down into three core challenges.
% \begin{enumerate}
%     \item \textbf{Mode Collapse}: Can the model generate animations with gestures in different modes of the underlying distribution? Does it sample poses from the correct mode?
%     \item \textbf{More than one gesture can be correct}: As more than one gesture can be correct for a given language and speech segment, metrics like L1 Loss and Probability of Correct Keypoints (PCK) \cite{simon2017hand} do not give the full picture. Can we relax the strictness of correct generation to define an objective metric that rewards for similar (\comm{or plausible}) gestures? 
%     \item \textbf{Language and Speech as predictors of gesture}: Gestures accompany spoken monologues or dialogues, but how good are language and speech as predictors of gestures?
% \end{enumerate}
%In this section, we explain the experiment setup which is used to evaluate these challenges of gesture generation from speech and language. 

% \comm{Talk about the split, individual level style and attribute level style, check out, maybe a subsection for that. also all multi-speaker models were trained for the same number of iterations}
% \comm{list of individual level styles and sttributes}
Our experiments are divided into 2 sections, (1) \textbf{Style Preservation:} Generating co-speech gestures for multiple speakers with their own individualistic style, (2) \textbf{Style Transfer:} Generating co-speech gestures with content (or audio) of a speaker and gesture style of another speaker. Additionally, style transfer can be speaker-level as well as attribute-level. We choose visually distinguishable attribute-level styles: (1) body orientation, (2) gesture frequency, (3) primary arm function and (4) sitting/standing posture. %These are listed in Table \ref{fig:attrtable}.
% All the experiments are conducted on the \datasetshortps introduced in Section \ref{sec:dataset}. \
% \comm{We train with 2 speakers 4 speakers and 8 speakers randomly chosen from the dataset. Also in a second study we take speakers with different attributes including frequency of gestures, orientation, ...}
%We start by briefly discussing the implementation details, which is followed by all the baseline models. 
We start by describing the baseline models followed by the evaluation metrics, which we will use to compare our model. We end this section with the description of our proposed dataset. %and a list of the speaker-level and attribute-level styles that were used for the experiments.%\footnote{We highly recommend looking at the generated videos in the supplementary for qualitative judgement.}.

\subsection{Baseline Models}
\subsubsection{\underline{Single-Speaker Models}}
: These models are not designed to perform style transfer and hence are not included for those experiments.
\begin{itemize}
    \item \noindent\textbf{Speech2Gesture \cite{ginosar2019learning}}: The closest work to co-speech gesture generation is one that only generates individualistic styles. We use the pre-trained models available from their code-base to render the videos. For rest of the speakers in \datasetshortss, we replicate their model, hyper-parameters and train speaker specific models. %We use these rendered videos for the human study and objective evaluation.
    \item \noindent\textbf{CMix-GAN }(variant of our model): As an ablation, we remove the style embedding module and style consistency losses from our model \modelshorts. Hence, a separate model is required to be trained for each speaker for style preservation experiments.
\end{itemize}

\subsubsection{\underline{Multi-speaker Models}}
\begin{itemize}
    \item \noindent\textbf{MUNIT \cite{huang2018multimodal}}: The closest work to our style-transfer task is MUNIT which takes multiple domains of images (i.e. uni-modal). We modify the encoders and decoders to domain specific architectures (i.e. 1D convolutions for audio instead of 2D convolutions for images) while retaining the loss functions.
    \item \noindent\textbf{StAGE }(variant of our model): As an ablation, we fix the number of sub-generators in our model \modelshorts to one. This is equivalent to setting $M=1$ in equation \ref{eq:mm_func}.
\end{itemize}

\subsection{Evaluation Metrics}
\subsubsection{Human Perceptual Study:} 
\label{ssec:human_study}
We conduct a human perceptual study on Amazon Mechanical Turk (AMT) for co-speech gesture generation (or style preservation) and style transfer (speaker-level and attribute-level) and measure preferences in two aspects of the generated animations, (1) \textbf{naturalness}, and (2) \textbf{style transfer correctness} for animation generation with content (i.e. audio) of speaker A and style of speaker B. We show a pair of videos with skeletal animations to the annotators. One of the animations is from the ground-truth set, while the other is generated using our proposed model. The generated animation could either have the same style or a different style as the original speaker. With unlimited time, the annotator has to answer two questions, (1) Which of the videos has more natural gestures? and (2) Do these videos have the same attribute-level style (or speaker-level style)?  The first question is a real vs. fake perceptual study against the ground truth, while the second question measures how often the algorithm is able to visually preserve or transfer style (attribute or individual level). We run this study for randomly selected 100 pairs of videos from the held-out set. .%\footnote{Some speakers did not have 100 videos in the test set. We used all the test videos in that case}  

\subsubsection{Probability of Correct Keypoints (PCK):} To measure the accuracy of the gesture generation, PCK \cite{andriluka20142d,simon2017hand} is used to evaluate all models. %If a predicted keypoint lies inside $\alpha \max(h, w)$ pixels around the target keypoint, then the prediction is deemed correct. $h$ and $w$ are height and width of the bounding box around the person. 
PCK values are averaged over $\alpha=0.1, 0.2$ as suggested in \cite{ginosar2019learning}.

\subsubsection{Mode Classification F1:}
Correctness of shape of a gesture can be quantified by measuring the number of times the model has sampled from the correct mode of the pose distribution. Formally, we use the true ($\textbf{Y}_p$) and generated ($\mathbf{\hat{Y}}^p$) pose to find the closest cluster $\hat{m}$ and $m$ respectively. If $m = \hat{m}$, the generated pose was sampled from the correct mode. F1 score of this $M$-class classification problem is defined as Mode Classification F1, or simply F1.%\footnote{We choose $M=8$ which is shown to yield the least variance in the F1 scores among values of $M$ from 2 to 12 and these results are shown in the supplementary.}  

\subsubsection{Inception Score (IS): } Generated pose sequences with the audio of speaker A and style of speaker B does not have a ground truth reference. To quantitatively measure the correctness and diversity of generated pose sequence we use the inception score \cite{salimans2016improved}. 
For generative tasks such as image generation, this metric has been used with a pre-trained classification network such as Inception Model \cite{szegedy2016rethinking}. In our case, the generated samples are not images, but a set of 2D keypoints. Hence, we train a network which classifies a sequence of poses to its corresponding speaker which estimates the conditional likelihood to calculate IS scores.
%This metric has been used extensively in research involving image generation and requires a pre-trained classification network (e.g. Inception Model \cite{szegedy2016rethinking}) to estimate $p(y|x)$ and $p(y)$, where $x$ is the generated image and $y$ is the class of the image. In our case, $x$ is the generated co-speech gestures and $y$ is the style (i.e. speaker identity). We train a network which classifies a sequence of poses to its corresponding speaker and use it to get IS scores for the generated pose sequences.
%\input{z72_attrtable}
%\vspace{-0.3cm}
\begin{figure}[ht]
\centering
\begin{subfigure}[t]{0.48\textwidth}
\includegraphics[width=\linewidth,trim={0 1cm 0 1cm},clip]{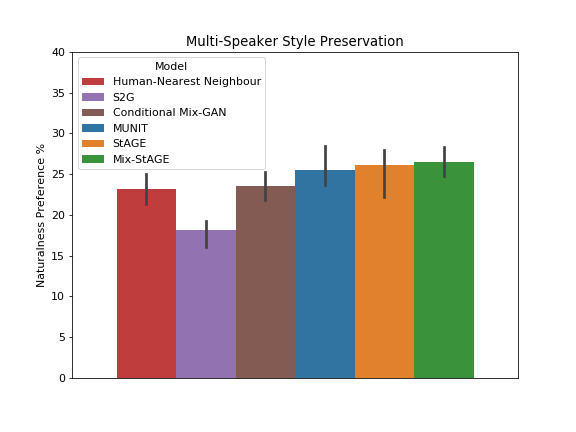}
\caption{Style Preservation Naturalness}
\end{subfigure}~
\begin{subfigure}[t]{0.25\textwidth}
\includegraphics[width=\linewidth,trim={0 0.4cm 0 0.5cm},clip]{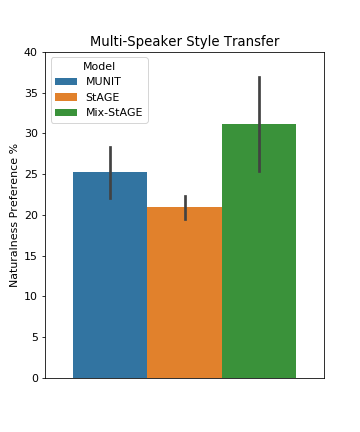}
\caption{Style Transfer Naturalness}
\end{subfigure}~
\begin{subfigure}[t]{0.25\textwidth}
\includegraphics[width=\linewidth,trim={0 0.4cm 0 0.5cm},clip]{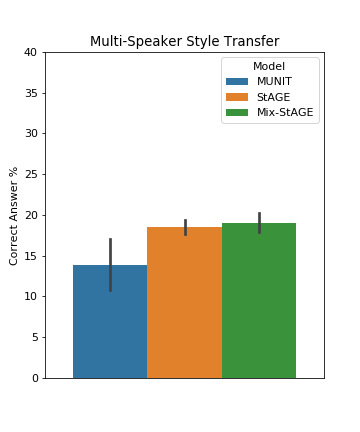}
\caption{Style Transfer Correctness}
\end{subfigure}
\caption{Perceptual Study for speaker-level style preservation in (a) and speaker level style transfer in (b), (c). We have naturalness preference for both style transfer and preservation, and style transfer correctness scores for style transfer. Higher is better. Error bars calculated for $p<0.1$ using a bootstrapped two sided t-test.}
\label{fig:human_study2}
\end{figure}

\subsection{\datasets}
\begin{tikzpicture}[remember picture,overlay]
\node[anchor=east,inner sep=4.0cm] at (current page text area.east|-0,1.2cm) {\includegraphics[width=0.8cm]{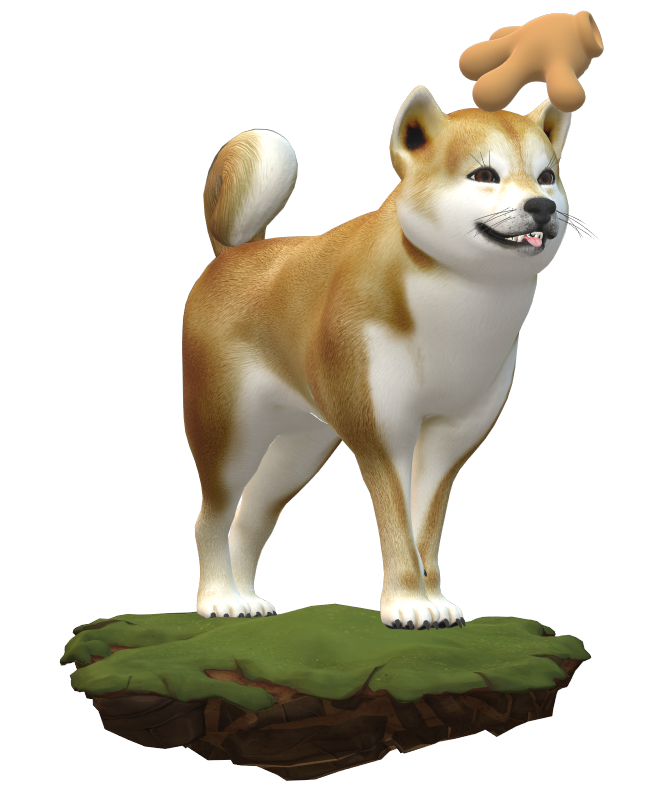}};
\end{tikzpicture}
Gesture styles, which may be defined by attributes such as type, frequency, orientation of the body, is representative of the idiosyncrasies of the speaker \cite{davis2019sometimes}. We create a new dataset, \datasetss, to study various styles of gestures for a large number of speakers in diverse settings. 

\datasetshortsss contains pose sequences aligned with corresponding audio signals and transcripts\footnote{While transcripts are a part of this dataset, they are ignored for the purposed of this work.} for 25 speakers (including 10 speakers from \cite{ginosar2019learning}) to offer a total of 251 hours of data, with a mean of 10.7 seconds and a standard deviation of 13.5 seconds per interval. The demographics of the speakers include 15 talk show hosts, 5 lecturers, 3 YouTubers, and 2 televangelists.

Each speaker's pose is represented via skeletal keypoints collected via OpenPose \cite{cao2018openpose} similar to \cite{ginosar2019learning}. It consists of of 52 coordinates of an individual's major joints for each frame at 15 frames per second, which we rescale by holding the length of each individual's shoulder constant. This prevents the model from encoding limb length in the style embeddings. Following prior work \cite{kucherenko2019analyzing,ginosar2019learning}, we represent audio features as mel spetrograms, which is a rich input representation shown to be useful for gesture generation. 

\begin{table}[]
\begin{tabular}{@{}l|c|c|c|c|c|c|c@{}}
\toprule
\multicolumn{1}{c|}{\multirow{2}{*}{\textbf{Model}}} & \multicolumn{3}{c|}{\textbf{Number of Speakers}}                & \multicolumn{4}{c|}{\textbf{Attributes}}                                                                                                                                                                                                                                                     \\ \cmidrule(l){2-8} 
\multicolumn{1}{c|}{}                                & \textbf{2 Speakers} & \textbf{4 Speakers} & \textbf{8 Speakers} & \textbf{\begin{tabular}[c]{@{}c@{}}Sitting vs\\ Standing\end{tabular}} & \textbf{\begin{tabular}[c]{@{}c@{}}Gesture \\ Frequency\end{tabular}} & \textbf{\begin{tabular}[c]{@{}c@{}}Body \\ Orientation\end{tabular}} & \textbf{\begin{tabular}[c]{@{}c@{}}Primary\\ Arm Func.\end{tabular}} \\ \midrule
\textbf{MUNIT} \cite{huang2018multimodal}                                      & 1.11                & 1.90                & 2.06                & 1.10                                                                   & 2.49                                                                  & 1.05                                                                 & 3.32                                                                 \\
\textbf{StAGE}                                       & 2.17                & \textbf{2.85}       & 3.89                & 1.68                                                                   & 4.38                                                                  & \textbf{6.81}                                                        & 3.14                                                                 \\
\textbf{Mix-StAGE}                                   & \textbf{2.61}       & \textbf{2.85}       & \textbf{4.48}       & \textbf{3.08}                                                          & \textbf{4.50}                                                         & 6.69                                                                 & \textbf{3.32}                                                        \\ \bottomrule
\end{tabular}
\caption{\textbf{Style Transfer}: Inception scores for style transfer on multi-speaker models (indicated in each row). Columns on the left refer to the speaker-level style transfer task while those on the right refer to the specific attribute-level style task. Bold numbers indicate $p<0.1$ in a bootstrapped two sided t-test.}
\label{tab:inception}
%\vspace{-3em}
\end{table}

\section{Results and Discussion}
\label{sec:results}
We group our results and discussions in (1) a first set of experiments studying style preservation (when output gesture styles are the same the original speaker) and (2) a second set of experiments studying transfer of gesture styles. 

\subsection{Gesture Animation and Style Preservation}
To understand the impact of adding more speakers, we select a random sample of 8 speakers for the largest 8-speaker multi-speaker model, and train smaller 4-speaker and 2-speaker models where the speakers trained are always a subset of the speakers that were trained in a larger model. This allows to compare the performance on the same two initial speakers which are `\textit{Corden}' and `\textit{lec\_cosmic}' in our case\footnote{The complete set of speakers used in our experiments are listed in the supplementary.}. We also compare with single-speaker models trained and tested on one speaker at a time. 

\subsubsection{Impact of training with Multiple Speakers}
%We train all multi-speaker models with 2, 4, and 8 speakers. 
Results from Table \ref{tab:main} show that multi-speaker models outperform single-speaker models especially for pose accuracy (i.e. PCK), shape and timing (i.e. F1). We find that increasing the number of speakers could sometimes reduce the performance of individual speakers but the overall performance generally shows improvement. 

\subsubsection{Comparison with previous baselines}
To compare with prior baselines, we focus first on the subjective evaluation shown in Figure \ref{fig:human_study2}\makered{a}, since it is arguably the most important metric. The results show consistent improvements on the naturalness rating for our proposed model \modelshorts and also our single-speaker variant CMix-GAN over the previous state of the art approach S2G \cite{ginosar2019learning}. We also observe that multi-speaker models perform better than single speaker-models. In Table \ref{tab:main}, we show similar quantitative improvements of \modelshorts and CMix-GAN over S2G for both PCK and F1 scores.
\begin{figure}[ht]
\centering
\begin{subfigure}[t]{0.48\textwidth}
\includegraphics[width=\linewidth,trim={0 0.2cm 0 0},clip]{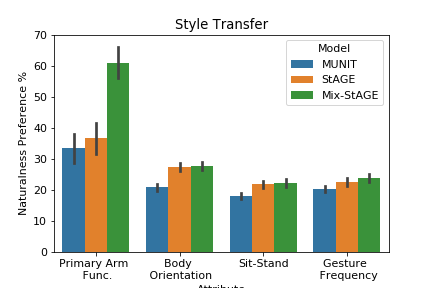}
\caption{Naturalness Preference}
\end{subfigure}~
\begin{subfigure}[t]{0.48\textwidth}
\includegraphics[width=\linewidth,trim={0 0.2cm 0 0},clip]{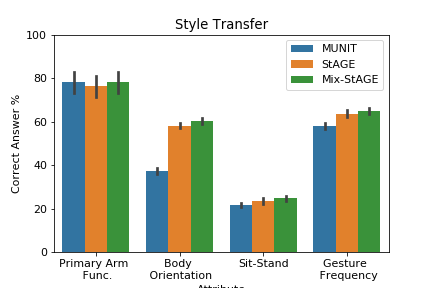}
\caption{Style Transfer Correctness}
\end{subfigure}
\caption{A visualization of the perceptual human study for attribute-level style transfer with (a) naturalness preference, and (b) style transfer correctness scores for the generated animations for a different style than the speaker. Higher is better. Error bars calculated for $p<0.1$ using a bootstrapped two sided t-test.}
\label{fig:human_study1}
\end{figure}

\subsubsection{Impact of Multiple Generators for Decoding}
\modelshort's gesture space models multiple modes, as seen in Figure \ref{fig:tsne}. Its importance is shown in Table \ref{tab:main} where models with single generators as the decoder (i.e. S2G, MUNIT and StAGE) showed lower F1 scores, most likely due to mode collapse while training. Multiple generators in CMix-GAN and Mix-StAGE boost F1 scores as compared to other models in the single-speaker and multi-speaker regimes respectively. A similar trend was observed in the perceptual study in Figure \ref{fig:human_study2}.

We also study the impact of the number of generators (hyperparameter M) in our \modelshorts model. While for small number of speakers (i.e. 2 speakers) a single generator is good enough, the positive effect of multiple generators can be observed as the number of speakers increase (see Table \ref{tab:main}). We also vary  $M \in \lbrace 1, 2, 4, 8, 12 \rbrace$ and observe that improvements seem to plateau at $M=8$ with only marginal improvements for larger number of sub-generators. For the ablation study we refer the readers to the supplementary. %M is a hyperparameter. Thus, it will naturally adapt in situations where more complexity is required.
% Choice of $M$ is a hyper-parameter and could vary depending on the task at hand. For gesture generation we conducted an ablation study for values of $M \in \lbrace 1, 2, 4, 8, 12 \rbrace$ which indicated that values $M>8$ the improvements in performance were not significant enough to warrant an increase in the number of generators. In situations with a gesture space with much more variety, which could happen with a large number of speakers, value of $M$ could be further increased as per requirement.
\subsubsection{Attribute-level Style Preservation in Multi-Speaker Models}
\begin{figure}[ht]
\centering
\begin{subfigure}[t]{0.48\textwidth}
\includegraphics[width=\linewidth,trim={0 0.2cm 0 0},clip]{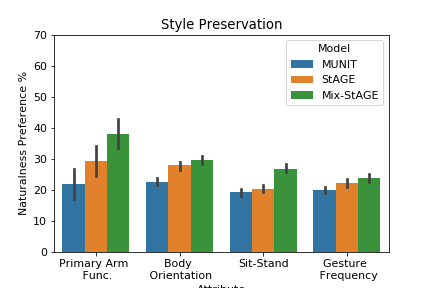}
\caption{Naturalness Preference}
\end{subfigure}~
\begin{subfigure}[t]{0.48\textwidth}
\includegraphics[width=\linewidth,trim={0 0.2cm 0 0},clip]{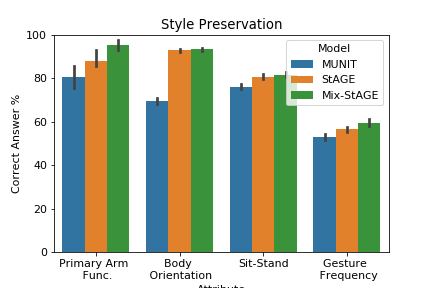}
\caption{Style Preservation Correctness}
\end{subfigure}
\caption{A visualization of the perceptual human study for attribute-level style preservation with (a) naturalness preference, and (b) style preservation correctness scores for the generated animations for the same style as the speaker. Higher is better. Error bars calculated for $p<0.1$ using a bootstrapped two sided t-test.}
\label{fig:human_study3}
\end{figure}

We also study style preservation for attributes in Section \ref{sec:exp} as a perceptual study in Figure \ref{fig:human_study3}. We observe that humans deem animations generated by \modelshorts  significantly more natural in most cases.  High scores ranging 60-90\% for style preservation correctness, with \modelshorts outperforming others, are observed for pairs of speakers in Figure \ref{fig:human_study3}\makered{b}. This indicates that style preservation may be a relatively easy task as compared to style transfer for multi-speaker models. With this, we now shift our focus to style transfer.

%\subsection{Preservation of Style within Multi-Speaker models}

%\vspace{-1cm}
\subsection{Style Transfer}
\subsubsection{Speaker-level Style Transfer}
To study our capability to transfer style of a specific speaker to a new speaker, we will compare the gesture spaces between the original speakers and the transferred speakers. Figure \ref{fig:tsne}\makered{a} shows that each original speaker occupies different regions in the $M^2$ gesture space. Using our \modelshorts model to transfer style, we can see the new gesture space in Figure \ref{fig:tsne}\makered{b}. For the transferred speakers the 2 spaces look quite similar. For instance, `\textit{Corden}' style (a speaker in our dataset) is represented by the color blue in Figure \ref{fig:tsne}\makered{a} and occupies the lower region of the gesture space. When \modelshorts generates co-speech gestures using audio of `\textit{Oliver}' and the style of `\textit{Corden}', it occupies a subset of `\textit{Corden's}' region in the gesture space, also represented by blue in Figure \ref{fig:tsne}\makered{b}. We see a similar trend for styles of `\textit{Oliver}` and `\textit{ytch\_prof}'. This is an indication of a successful style transfer across different speakers. We note the lack of clean separation in the gesture space among different styles as there could common gestures across multiple speakers.

For the perceptual study, we want to know if humans can distinguish the generated speaker styles. For this, we show human annotators two videos: a ground truth video in a specific style, and a generated video which is either from the style of the same speaker or a different speaker. Annotators have to decide if this is the same style or not. We use the 4-speaker model for this experiment. Figure \ref{fig:human_study2}\makered{b} shows naturalness preference and \ref{fig:human_study2}\makered{c} shows percentage of the time style was transferred correctly. Our model \modelshorts performs best in both cases. This trend is corroborated with higher inception scores in Table \ref{tab:inception}. 

%one of which is the ground truth and the other is an animation generated by an algorithm - our model or a baseline. The generated animation could either have the same style of the original speaker, or of a different speaker. The annotators have to judge whether the style of gestures in both the videos is the same.

\subsubsection{Impact of Number of Speakers for Style Transfer}
In Table \ref{tab:inception}, we observe that increasing the number of speakers used for training also increases the average inception score for the stylized gesture generations. This is a welcome effect as it indicates increases in the diversity and the accuracy of the generations.

%\ref{fig:tsne_pair}
\begin{figure}[h]
\centering
\begin{subfigure}[t]{0.235\textwidth}
\includegraphics[width=\linewidth]{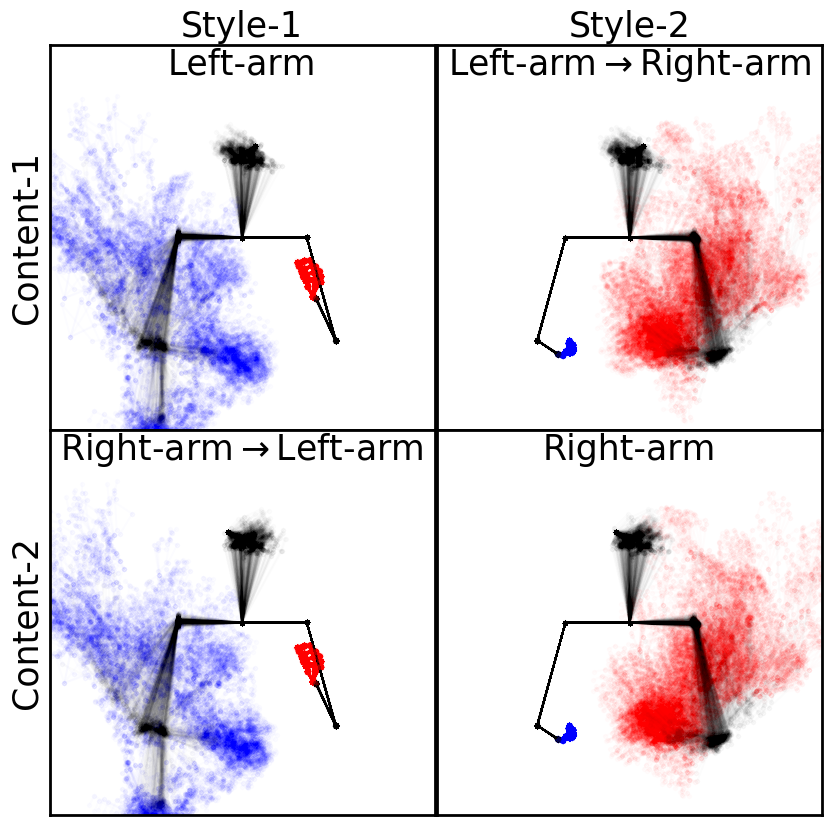}
\caption{Primary Arm Func.}
\end{subfigure}
~
\begin{subfigure}[t]{0.235\textwidth}
\includegraphics[width=\linewidth]{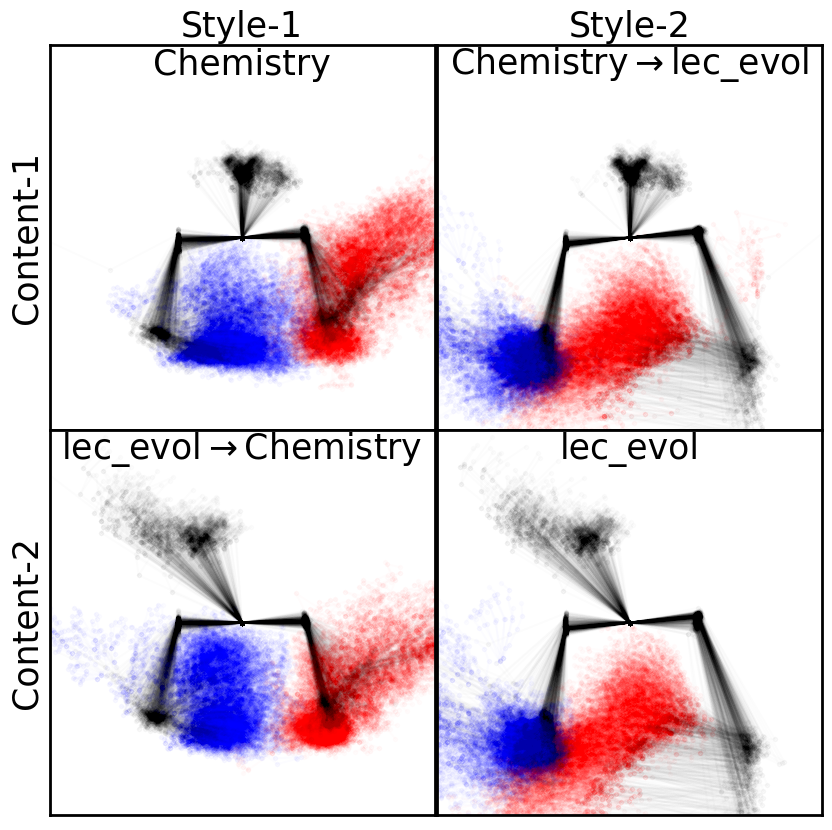}
\caption{Body Orientation}
\end{subfigure}
~
\begin{subfigure}[t]{0.235\textwidth}
\includegraphics[width=\linewidth]{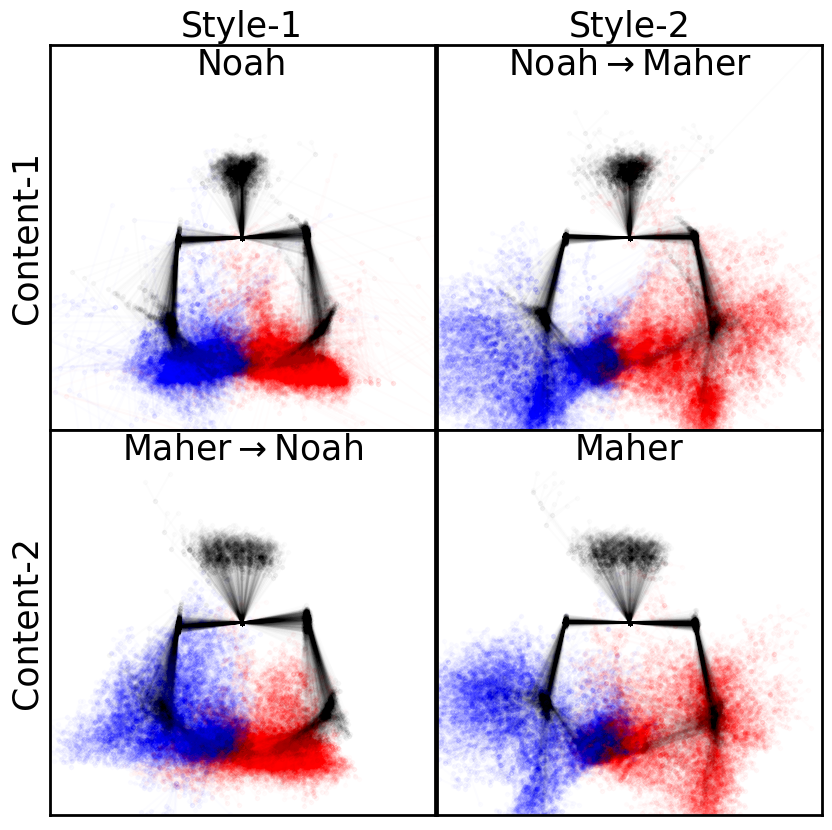}
\caption{Sitting vs Standing}
\end{subfigure}
~
\begin{subfigure}[t]{0.235\textwidth}
\includegraphics[width=\linewidth]{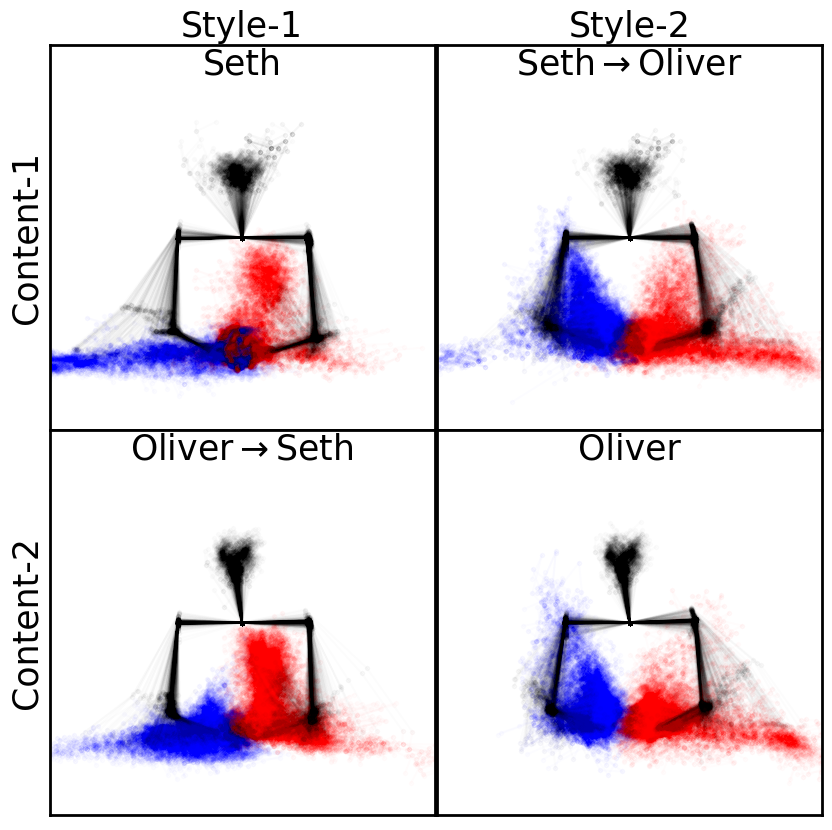}
\caption{Gesture Frequency}
\end{subfigure}
\caption{Style-Content Heatmaps for attribute-level style transfer. Each column represents the same style, while rows have input audio from different speakers. These heatmaps show that gestures are consistent across audio inputs but different between styles. Red regions correspond to the motion of the right arm, while blue corresponds to the left.}
\label{fig:att_heatmap}
\end{figure}
%\vspace{-1cm}
\subsubsection{Attribute-level Style Transfer in Multi-Speaker Models}
We study four common attributes of gesture style which are also visually distinguishable by humans: (1) sitting vs. standing, (2) high vs low gesture frequency, (3) left vs right body orientation and (4) left vs right primary arm. Speakers were selected carefully to represent each extremes of these four attributes. We run a perceptual study similar to the one for speaker-level styles. However, we ask the annotators to judge if the attribute is the same in both of the videos (e.g. are both the people gesturing with the same arm?). Results from Figure \ref{fig:human_study1} show that \modelshorts generates more (or similar) number of natural gestures with the correct attribute-level style compared to the other baselines. We also observe that it is harder for humans to determine if a person is standing or sitting, which we suspect is due to the missing waistline in the animation.

For a visual understanding of the generated gestures and stylized gestures, we plot a style-content heatmap in Figure \ref{fig:att_heatmap}, where columns represent generations for a specific style, while rows represent different speaker's audio as input. These heatmaps show that gestures are consistent across audio inputs but different between styles. Accuracy and diversity of style transfer is corroborated by inception scores in Table \ref{tab:inception}.

\section{Conclusions}
\label{sec:conclusions}
In this paper, we propose a new model, named \modelshort, which learns a single model for multiple speakers while learning unique style embeddings for each speaker's gestures in an end-to-end manner. A novelty of \modelshorts was to learn a mixture of generative models conditioned on gesture style while the audio drives the co-speech gesture generation. We also introduced a new dataset, \datasetss, designed to study gesture generation and style transfer. It consists of 25 speakers (15 new speakers and 10 speakers from Ginosar et. al. \cite{ginosar2019learning}) for a total of 250+ hours of gestures and aligned audio signals. Our proposed \modelshorts model significantly outperformed previous state-of-the-art approach for gesture generation and provided a path towards performing gesture style transfer across multiple speakers. We also demonstrated, through human perceptual studies, that the generated animations by our model are more natural whilst being able to retain or transfer style.
\subsubsection{Acknowledgements}
This material is based upon work partially supported by the National Science Foundation (Awards \#1750439 \#1722822), National Institutes of Health and the InMind project. Any opinions, findings, and conclusions or recommendations expressed in this material are those of the author(s) and do not necessarily reflect the views of National Science Foundation or National Institutes of Health, and no official endorsement should be inferred.

% \section{Appendix}
% \label{sec:appendix}
% \input{9_appendix.tex}

%
\bibliographystyle{splncs04}
\bibliography{egbib}

% {\small
% \bibliographystyle{ieee_fullname}
% \bibliography{egbib}
% }

%\newpage
\begin{center}
    \textbf{{\Large \underline{Supplementary}}}
\end{center}
\appendix
\section{\datasetshorts}
\subsection{Speaker List}
\begin{figure*}[h]
\includegraphics[width=\linewidth, trim={0 0.6cm 0 0.3cm},clip]{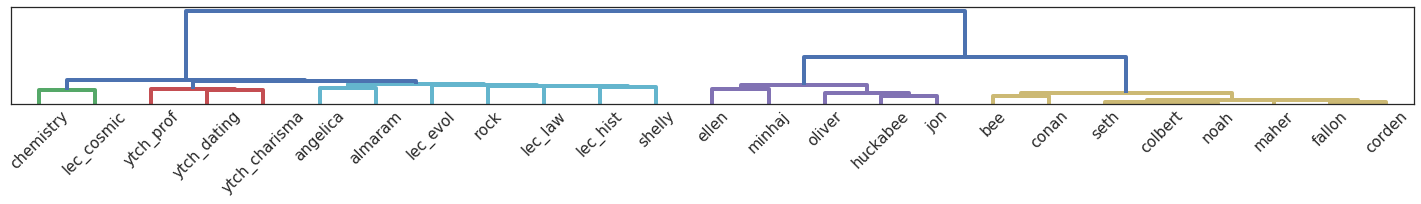}
\caption{List of speakers in the dataset as a dendrogram based on the content of the speech.} \label{fig:dendrogram}
\end{figure*}

The list of speakers in the dataset are in Figure \ref{fig:dendrogram} as a dendrogram. This dendogram was created using text as the discriminating features. Speakers within the same cluster have a similar vocabulary. For the purposes of our experiments we use the speakers listed in Table \ref{tab:evil} and \ref{tab:main_complete}.

\subsection{Attributes}
We define 4 different attributes in Table \ref{fig:attrtable} and select pairs of speakers that demonstrate visually striking differences with respect to those attributes.

% \usepackage{booktabs}
\begin{table}[]
\centering
\begin{tabular}{@{}l|l|l|l|l|l|l|l@{}}
\toprule
\multicolumn{2}{c|}{\textbf{Sitting/Standing}} & \multicolumn{2}{c|}{\textbf{Gesture Frequency}} & \multicolumn{2}{c|}{\textbf{Body Orientation}} & \multicolumn{2}{c|}{\textbf{Primary Arm Func.}}      \\ \midrule
\multicolumn{2}{l|}{\textbf{Sitting: }Noah}    & \multicolumn{2}{l|}{\textbf{Low: } Seth}         & \multicolumn{2}{l|}{\textbf{Right: } Chemistry} & \multicolumn{2}{l|}{\textbf{Right Arm: } lec\_cosmic} \\
\multicolumn{2}{l|}{\textbf{Standing: } Maher}  & \multicolumn{2}{l|}{\textbf{High: } Oliver}      & \multicolumn{2}{l|}{\textbf{Left: } Oliver}     & \multicolumn{2}{l|}{\textbf{Left Arm: } lec\_cosmic}  \\ \bottomrule
\end{tabular}
\caption{Selection of speakers for attribute-level style modeling }
\label{fig:attrtable}
\vspace{-3.5em}
\end{table}

\section{Other Results, Discussions and Future Directions}
These results complement Section 6 of the main paper with a few more observations, explorations and ablation studies. This is followed by potential future directions.
\subsubsection{Attribute-Level Style Preservation}
Gesture generation for pairs of speakers shows improvements in PCK and F1 scores (see Table \ref{tab:evil}) following the trend of perceptual study in Figure 6 of the main paper.
% \usepackage{booktabs}
% \usepackage{multirow}
\begin{table}[]
\centering
\begin{tabular}{l|l|rr|rr|rr|rr|rr}
\hline
\multicolumn{1}{c|}{}                                      & \multicolumn{1}{c|}{}                                    & \multicolumn{4}{c|}{\textbf{Single-Speaker Models}}                                                                                                                                                                   & \multicolumn{6}{c|}{\textbf{Multi-Speaker Models}}                                                                                                                                                                                                                                                                                \\ \cline{3-12} 
\multicolumn{1}{c|}{}                                      & \multicolumn{1}{c|}{}                                    & \multicolumn{2}{c|}{\textbf{S2G}}                                                                         & \multicolumn{2}{c|}{\textbf{CMix-GAN}}                                                                    & \multicolumn{2}{c|}{\textbf{MUNIT}}                                                                       & \multicolumn{2}{c|}{\textbf{StAGE}}                                                                       & \multicolumn{2}{c|}{\textbf{Mix-StAGE}}                                                                   \\ \cline{3-12} 
\multicolumn{1}{c|}{\multirow{-3}{*}{\textbf{Attributes}}} & \multicolumn{1}{c|}{\multirow{-3}{*}{\textbf{Speakers}}} & \multicolumn{1}{c}{\textbf{PCK}}                    & \multicolumn{1}{c|}{\textbf{F1}}                    & \multicolumn{1}{c}{\textbf{PCK}}                    & \multicolumn{1}{c|}{\textbf{F1}}                    & \multicolumn{1}{c}{\textbf{PCK}}                    & \multicolumn{1}{c|}{\textbf{F1}}                    & \multicolumn{1}{c}{\textbf{PCK}}                    & \multicolumn{1}{c|}{\textbf{F1}}                    & \multicolumn{1}{c}{\textbf{PCK}}                    & \multicolumn{1}{c}{\textbf{F1}}                     \\ \hline
\rowcolor[HTML]{FFFFFF} 
\cellcolor[HTML]{EFEFEF}\textbf{Sitting/Standing}          & \cellcolor[HTML]{EFEFEF}\textbf{Mean}                    & {\color[HTML]{111111} 0.35}                         & {\color[HTML]{111111} 0.12}                         & {\color[HTML]{111111} 0.35}                         & {\color[HTML]{111111} 0.25}                         & {\color[HTML]{111111} 0.36}                         & {\color[HTML]{111111} 0.07}                         & {\color[HTML]{111111} \textbf{0.42}}                & {\color[HTML]{111111} 0.18}                         & {\color[HTML]{111111} \textbf{0.42}}                & {\color[HTML]{111111} \textbf{0.25}}                \\
Sitting                                                    & Noah                                                     & \cellcolor[HTML]{FFFFFF}{\color[HTML]{111111} 0.45} & \cellcolor[HTML]{FFFFFF}{\color[HTML]{111111} 0.11} & \cellcolor[HTML]{FFFFFF}{\color[HTML]{111111} 0.45} & \cellcolor[HTML]{FFFFFF}{\color[HTML]{111111} 0.28} & \cellcolor[HTML]{FFFFFF}{\color[HTML]{111111} 0.34} & \cellcolor[HTML]{FFFFFF}{\color[HTML]{111111} 0.09} & \cellcolor[HTML]{FFFFFF}{\color[HTML]{111111} 0.44} & \cellcolor[HTML]{FFFFFF}{\color[HTML]{111111} 0.14} & \cellcolor[HTML]{FFFFFF}{\color[HTML]{111111} 0.44} & \cellcolor[HTML]{FFFFFF}{\color[HTML]{111111} 0.26} \\
Standing                                                   & Maher                                                    & \cellcolor[HTML]{FFFFFF}{\color[HTML]{111111} 0.25} & \cellcolor[HTML]{FFFFFF}{\color[HTML]{111111} 0.13} & \cellcolor[HTML]{FFFFFF}{\color[HTML]{111111} 0.24} & \cellcolor[HTML]{FFFFFF}{\color[HTML]{111111} 0.22} & \cellcolor[HTML]{FFFFFF}{\color[HTML]{111111} 0.22} & \cellcolor[HTML]{FFFFFF}{\color[HTML]{111111} 0.07} & \cellcolor[HTML]{FFFFFF}{\color[HTML]{111111} 0.28} & \cellcolor[HTML]{FFFFFF}{\color[HTML]{111111} 0.26} & \cellcolor[HTML]{FFFFFF}{\color[HTML]{111111} 0.26} & \cellcolor[HTML]{FFFFFF}{\color[HTML]{111111} 0.25} \\ \hline
\rowcolor[HTML]{EFEFEF} 
\textbf{Gesture Frequency}                                 & \textbf{Mean}                                            & 0.55                                                & 0.41                                                & 0.56                                                & 0.44                                                & 0.34                                                & 0.14                                                & \textbf{0.58}                                       & 0.51                                                & \textbf{0.58}                                       & \textbf{0.53}                                       \\
Low                                                        & Seth                                                     & 0.56                                                & 0.50                                                & 0.58                                                & 0.54                                                & 0.22                                                & 0.02                                                & 0.58                                                & 0.54                                                & 0.59                                                & 0.57                                                \\
High                                                       & Oliver                                                   & 0.54                                                & 0.32                                                & 0.54                                                & 0.34                                                & 0.35                                                & 0.22                                                & 0.54                                                & 0.38                                                & 0.56                                                & 0.42                                                \\ \hline
\rowcolor[HTML]{EFEFEF} 
\textbf{Body Orientation}                                  & \textbf{Mean}                                            & 0.39                                                & 0.14                                                & \textbf{0.43}                                       & 0.25                                                & 0.14                                                & 0.05                                                & 0.40                                                & \textbf{0.42}                                       & 0.40                                                & \textbf{0.40}                                       \\
Right                                                      & Chemistry                                                & 0.35                                                & 0.23                                                & 0.36                                                & 0.27                                                & 0.15                                                & 0.05                                                & 0.37                                                & 0.39                                                & 0.40                                                & 0.39                                                \\
Left                                                       & lec\_evol                                                & 0.44                                                & 0.05                                                & 0.50                                                & 0.23                                                & 0.28                                                & 0.34                                                & 0.50                                                & 0.44                                                & 0.49                                                & 0.46                                                \\ \hline
\rowcolor[HTML]{EFEFEF} 
\textbf{Primary Arm Func.}                                 & \textbf{Mean}                                            & 0.43                                                & 0.12                                                & 0.43                                                & 0.33                                                & 0.35                                                & 0.02                                                & 0.59                                                & 0.30                                                & \textbf{0.61}                                       & \textbf{0.37}                                       \\
Left Arm                                                   & lec\_cosmic                                              & 0.41                                                & 0.08                                                & 0.41                                                & 0.28                                                & 0.32                                                & 0.06                                                & 0.60                                                & 0.27                                                & 0.62                                                & 0.36                                                \\
Right Arm                                                  & lec\_cosmic                                              & 0.45                                                & 0.18                                                & 0.45                                                & 0.38                                                & 0.44                                                & 0.12                                                & 0.58                                                & 0.31                                                & 0.60                                                & 0.35                                                \\ \hline
\end{tabular}
\caption{Objective metrics for attribute-level style preservation of single-speaker and multi-speaker models as indicated in the columns. Each row refers to the number of speakers the model was trained, with the average performance indicated at the top. The scores for common individual speakers are also indicated below alongside. For detailed results on other speakers please refer to the supplementary}
\label{tab:evil}
\end{table}

\subsubsection{Speaker-Level Style Preservation}
Complete numerical results for speaker-level style preservation (for Table 1 in the main paper) are listed in Table \ref{tab:main_complete}. The PCK and F1 scores of the individual speakers show the same trend as the average score for each model.
% \usepackage{booktabs}
% \usepackage{multirow}
\begin{table}[]
\centering
\begin{tabular}{c|l|rr|rr|rr|rr|rr}
\hline
                                                                                      & \multicolumn{1}{c|}{}                                   & \multicolumn{4}{c|}{\textbf{Single-Speaker Models}}                                                                                            & \multicolumn{6}{c|}{\textbf{Multi-Speaker Models}}                                                                                                                                                                             \\ \cline{3-12} 
                                                                                      & \multicolumn{1}{c|}{}                                   & \multicolumn{2}{c|}{\textbf{S2G}}                                   & \multicolumn{2}{c|}{\textbf{CMix-GAN}}                                   & \multicolumn{2}{c|}{\textbf{MUNIT}}                                 & \multicolumn{2}{c|}{\textbf{StAGE}}                                      & \multicolumn{2}{c|}{\textbf{Mix-StAGE}}                                       \\ \cline{3-12} 
\multirow{-3}{*}{\textbf{\begin{tabular}[c]{@{}c@{}}No. of \\ Speakers\end{tabular}}} & \multicolumn{1}{c|}{\multirow{-3}{*}{\textbf{Speaker}}} & \multicolumn{1}{c}{\textbf{PCK}} & \multicolumn{1}{c|}{\textbf{F1}} & \multicolumn{1}{c}{\textbf{PCK}} & \multicolumn{1}{c|}{\textbf{F1}}      & \multicolumn{1}{c}{\textbf{PCK}} & \multicolumn{1}{c|}{\textbf{F1}} & \multicolumn{1}{c}{\textbf{PCK}} & \multicolumn{1}{c|}{\textbf{F1}}      & \multicolumn{1}{c}{\textbf{PCK}}      & \multicolumn{1}{c}{\textbf{F1}}       \\ \hline
                                                                                      & \cellcolor[HTML]{EFEFEF}\textbf{Mean}                   & \cellcolor[HTML]{EFEFEF}0.25     & \cellcolor[HTML]{EFEFEF}0.08     & \cellcolor[HTML]{EFEFEF}0.26     & \cellcolor[HTML]{EFEFEF}\textbf{0.27} & \cellcolor[HTML]{EFEFEF}0.24     & \cellcolor[HTML]{EFEFEF}0.06     & \cellcolor[HTML]{EFEFEF}0.36     & \cellcolor[HTML]{EFEFEF}0.21          & \cellcolor[HTML]{EFEFEF}\textbf{0.34} & \cellcolor[HTML]{EFEFEF}0.22          \\ \cline{2-12} 
                                                                                      & Corden                                                  & \cellcolor[HTML]{FFFFFF}0.30     & \cellcolor[HTML]{FFFFFF}0.05     & \cellcolor[HTML]{FFFFFF}0.32     & \cellcolor[HTML]{FFFFFF}0.21          & \cellcolor[HTML]{FFFFFF}0.25     & \cellcolor[HTML]{FFFFFF}0.06     & \cellcolor[HTML]{FFFFFF}0.36     & \cellcolor[HTML]{FFFFFF}0.21          & \cellcolor[HTML]{FFFFFF}0.34          & \cellcolor[HTML]{FFFFFF}0.22          \\
\multirow{-3}{*}{\textbf{2}}                                                          & lec\_cosmic                                             & \cellcolor[HTML]{FFFFFF}0.19     & \cellcolor[HTML]{FFFFFF}0.12     & \cellcolor[HTML]{FFFFFF}0.19     & \cellcolor[HTML]{FFFFFF}0.33          & \cellcolor[HTML]{FFFFFF}0.15     & \cellcolor[HTML]{FFFFFF}0.19     & \cellcolor[HTML]{FFFFFF}0.20     & \cellcolor[HTML]{FFFFFF}0.48          & \cellcolor[HTML]{FFFFFF}0.24          & \cellcolor[HTML]{FFFFFF}0.49          \\ \hline
                                                                                      & \cellcolor[HTML]{EFEFEF}\textbf{Mean}                   & \cellcolor[HTML]{EFEFEF}0.37     & \cellcolor[HTML]{EFEFEF}0.18     & \cellcolor[HTML]{EFEFEF}0.37     & \cellcolor[HTML]{EFEFEF}0.27          & \cellcolor[HTML]{EFEFEF}0.22     & \cellcolor[HTML]{EFEFEF}0.03     & \cellcolor[HTML]{EFEFEF}0.38     & \cellcolor[HTML]{EFEFEF}\textbf{0.34} & \cellcolor[HTML]{EFEFEF}\textbf{0.39} & \cellcolor[HTML]{EFEFEF}\textbf{0.35} \\ \cline{2-12} 
                                                                                      & Corden                                                  & \cellcolor[HTML]{FFFFFF}0.30     & \cellcolor[HTML]{FFFFFF}0.05     & \cellcolor[HTML]{FFFFFF}0.32     & \cellcolor[HTML]{FFFFFF}0.21          & \cellcolor[HTML]{FFFFFF}0.24     & \cellcolor[HTML]{FFFFFF}0.07     & \cellcolor[HTML]{FFFFFF}0.35     & \cellcolor[HTML]{FFFFFF}0.27          & \cellcolor[HTML]{FFFFFF}0.35          & \cellcolor[HTML]{FFFFFF}0.30          \\
                                                                                      & lec\_cosmic                                             & \cellcolor[HTML]{FFFFFF}0.19     & \cellcolor[HTML]{FFFFFF}0.12     & \cellcolor[HTML]{FFFFFF}0.19     & \cellcolor[HTML]{FFFFFF}0.33          & \cellcolor[HTML]{FFFFFF}0.19     & \cellcolor[HTML]{FFFFFF}0.16     & \cellcolor[HTML]{FFFFFF}0.18     & \cellcolor[HTML]{FFFFFF}0.23          & \cellcolor[HTML]{FFFFFF}0.20          & \cellcolor[HTML]{FFFFFF}0.19          \\
                                                                                      & ytch\_prof                                              & \cellcolor[HTML]{FFFFFF}0.43     & \cellcolor[HTML]{FFFFFF}0.22     & \cellcolor[HTML]{FFFFFF}0.43     & \cellcolor[HTML]{FFFFFF}0.22          & \cellcolor[HTML]{FFFFFF}0.15     & \cellcolor[HTML]{FFFFFF}0.02     & \cellcolor[HTML]{FFFFFF}0.42     & \cellcolor[HTML]{FFFFFF}0.34          & \cellcolor[HTML]{FFFFFF}0.40          & \cellcolor[HTML]{FFFFFF}0.32          \\
\multirow{-5}{*}{\textbf{4}}                                                          & Oliver                                                  & \cellcolor[HTML]{FFFFFF}0.54     & \cellcolor[HTML]{FFFFFF}0.32     & \cellcolor[HTML]{FFFFFF}0.54     & \cellcolor[HTML]{FFFFFF}0.34          & \cellcolor[HTML]{FFFFFF}0.20     & \cellcolor[HTML]{FFFFFF}0.09     & \cellcolor[HTML]{FFFFFF}0.54     & \cellcolor[HTML]{FFFFFF}0.47          & \cellcolor[HTML]{FFFFFF}0.55          & \cellcolor[HTML]{FFFFFF}0.52          \\ \hline
                                                                                      & \cellcolor[HTML]{EFEFEF}\textbf{Mean}                   & \cellcolor[HTML]{EFEFEF}0.36     & \cellcolor[HTML]{EFEFEF}0.14     & \cellcolor[HTML]{EFEFEF}0.37     & \cellcolor[HTML]{EFEFEF}0.26          & \cellcolor[HTML]{EFEFEF}0.31     & \cellcolor[HTML]{EFEFEF}0.15     & \cellcolor[HTML]{EFEFEF}0.38     & \cellcolor[HTML]{EFEFEF}\textbf{0.32} & \cellcolor[HTML]{EFEFEF}\textbf{0.40} & \cellcolor[HTML]{EFEFEF}\textbf{0.33} \\ \cline{2-12} 
                                                                                      & Corden                                                  & \cellcolor[HTML]{FFFFFF}0.30     & \cellcolor[HTML]{FFFFFF}0.05     & \cellcolor[HTML]{FFFFFF}0.32     & \cellcolor[HTML]{FFFFFF}0.21          & \cellcolor[HTML]{FFFFFF}0.23     & \cellcolor[HTML]{FFFFFF}0.03     & \cellcolor[HTML]{FFFFFF}0.32     & \cellcolor[HTML]{FFFFFF}0.28          & \cellcolor[HTML]{FFFFFF}0.36          & \cellcolor[HTML]{FFFFFF}0.27          \\
                                                                                      & lec\_cosmic                                             & \cellcolor[HTML]{FFFFFF}0.19     & \cellcolor[HTML]{FFFFFF}0.12     & \cellcolor[HTML]{FFFFFF}0.19     & \cellcolor[HTML]{FFFFFF}0.33          & \cellcolor[HTML]{FFFFFF}0.13     & \cellcolor[HTML]{FFFFFF}0.09     & \cellcolor[HTML]{FFFFFF}0.23     & \cellcolor[HTML]{FFFFFF}0.34          & \cellcolor[HTML]{FFFFFF}0.24          & \cellcolor[HTML]{FFFFFF}0.32          \\
                                                                                      & ytch\_prof                                              & \cellcolor[HTML]{FFFFFF}0.43     & \cellcolor[HTML]{FFFFFF}0.22     & \cellcolor[HTML]{FFFFFF}0.43     & \cellcolor[HTML]{FFFFFF}0.22          & \cellcolor[HTML]{FFFFFF}0.39     & \cellcolor[HTML]{FFFFFF}0.37     & \cellcolor[HTML]{FFFFFF}0.44     & \cellcolor[HTML]{FFFFFF}0.39          & \cellcolor[HTML]{FFFFFF}0.45          & \cellcolor[HTML]{FFFFFF}0.39          \\
                                                                                      & Oliver                                                  & \cellcolor[HTML]{FFFFFF}0.54     & \cellcolor[HTML]{FFFFFF}0.32     & \cellcolor[HTML]{FFFFFF}0.54     & \cellcolor[HTML]{FFFFFF}0.34          & \cellcolor[HTML]{FFFFFF}0.35     & \cellcolor[HTML]{FFFFFF}0.30     & \cellcolor[HTML]{FFFFFF}0.54     & \cellcolor[HTML]{FFFFFF}0.39          & \cellcolor[HTML]{FFFFFF}0.54          & \cellcolor[HTML]{FFFFFF}0.46          \\
                                                                                      & Ellen                                                   & \cellcolor[HTML]{FFFFFF}0.29     & \cellcolor[HTML]{FFFFFF}0.13     & \cellcolor[HTML]{FFFFFF}0.30     & \cellcolor[HTML]{FFFFFF}0.23          & \cellcolor[HTML]{FFFFFF}0.33     & \cellcolor[HTML]{FFFFFF}0.17     & \cellcolor[HTML]{FFFFFF}0.34     & \cellcolor[HTML]{FFFFFF}0.21          & \cellcolor[HTML]{FFFFFF}0.33          & \cellcolor[HTML]{FFFFFF}0.25          \\
                                                                                      & Noah                                                    & \cellcolor[HTML]{FFFFFF}0.45     & \cellcolor[HTML]{FFFFFF}0.11     & \cellcolor[HTML]{FFFFFF}0.45     & \cellcolor[HTML]{FFFFFF}0.28          & \cellcolor[HTML]{FFFFFF}0.40     & \cellcolor[HTML]{FFFFFF}0.23     & \cellcolor[HTML]{FFFFFF}0.44     & \cellcolor[HTML]{FFFFFF}0.24          & \cellcolor[HTML]{FFFFFF}0.44          & \cellcolor[HTML]{FFFFFF}0.27          \\
                                                                                      & lec\_evol                                               & \cellcolor[HTML]{FFFFFF}0.44     & \cellcolor[HTML]{FFFFFF}0.05     & \cellcolor[HTML]{FFFFFF}0.50     & \cellcolor[HTML]{FFFFFF}0.23          & \cellcolor[HTML]{FFFFFF}0.33     & \cellcolor[HTML]{FFFFFF}0.42     & \cellcolor[HTML]{FFFFFF}0.45     & \cellcolor[HTML]{FFFFFF}0.66          & \cellcolor[HTML]{FFFFFF}0.48          & \cellcolor[HTML]{FFFFFF}0.66          \\
\multirow{-9}{*}{\textbf{8}}                                                          & Maher                                                   & \cellcolor[HTML]{FFFFFF}0.25     & \cellcolor[HTML]{FFFFFF}0.13     & \cellcolor[HTML]{FFFFFF}0.24     & \cellcolor[HTML]{FFFFFF}0.22          & \cellcolor[HTML]{FFFFFF}0.23     & \cellcolor[HTML]{FFFFFF}0.17     & \cellcolor[HTML]{FFFFFF}0.25     & \cellcolor[HTML]{FFFFFF}0.25          & \cellcolor[HTML]{FFFFFF}0.25          & \cellcolor[HTML]{FFFFFF}0.25          \\ \hline
\end{tabular}
\caption{Objective metrics for speaker-level style preservation of single-speaker and multi-speaker models as indicated in the columns. Each row refers to the number of speakers the model was trained, with the average performance indicated at the top. The scores for individual speakers are also indicated below alongside. * refers to a Single-speaker Model}
\label{tab:main_complete}
\end{table}

\subsubsection{Impact of value of $M$ on gesture generation}
We run an ablation study on the choice of $M$ for the pose decoder. We report the average of PCK and F1 scores in Table \ref{tab:mode_ablation} which were calculated for each speaker in single-speaker models. We find the the scores plateau with increasing values of $M$ for single speaker models unlike multi-speaker models like \modelshorts.
\begin{table}[]
\centering
\begin{tabular}{l|c|c}
\hline
\multicolumn{1}{c|}{\multirow{2}{*}{\textbf{\begin{tabular}[c]{@{}c@{}}Single Speaker\\ Models\end{tabular}}}} & \multicolumn{2}{c|}{\textbf{Metrics}}            \\ \cline{2-3} 
\multicolumn{1}{c|}{}                                                                                          & \textbf{F1 $\uparrow$} & \textbf{PCK $\uparrow$} \\ \hline
\textbf{S2G}                                                                                                   & 18.9                   & 36.6                    \\
\textbf{CMix-GAN ($M=1$)}                                                                                      & 26.6                   & \textbf{37.9}           \\
\textbf{CMix-GAN ($M=4$)}                                                                                      & 27.7                   & 36.6                    \\
\textbf{CMix-GAN ($M=8$)}                                                                                      & \textbf{28.0}          & 36.7                    \\
\textbf{CMix-GAN ($M=12$)}                                                                                     & \textbf{27.8}          & \textbf{37.0}           \\ \hline
\end{tabular}
\caption{Comparision of \modelshorts with different values of $M$ over F1 and PCK. The results are reported as a mean over all speakers in \datasetshortss. We can see that the performance for single speaker models does not improve by increasing the number of modes $M$. This is unlike multi-speaker models, where the addition of sub-generators gives the model an edge over single-speaker models.}
\end{table}

\subsubsection{Exploring Style Control}
\begin{figure}%{l}{0.2\textwidth}
%\begin{figure}
\begin{center}
    \includegraphics[width=0.25\linewidth]{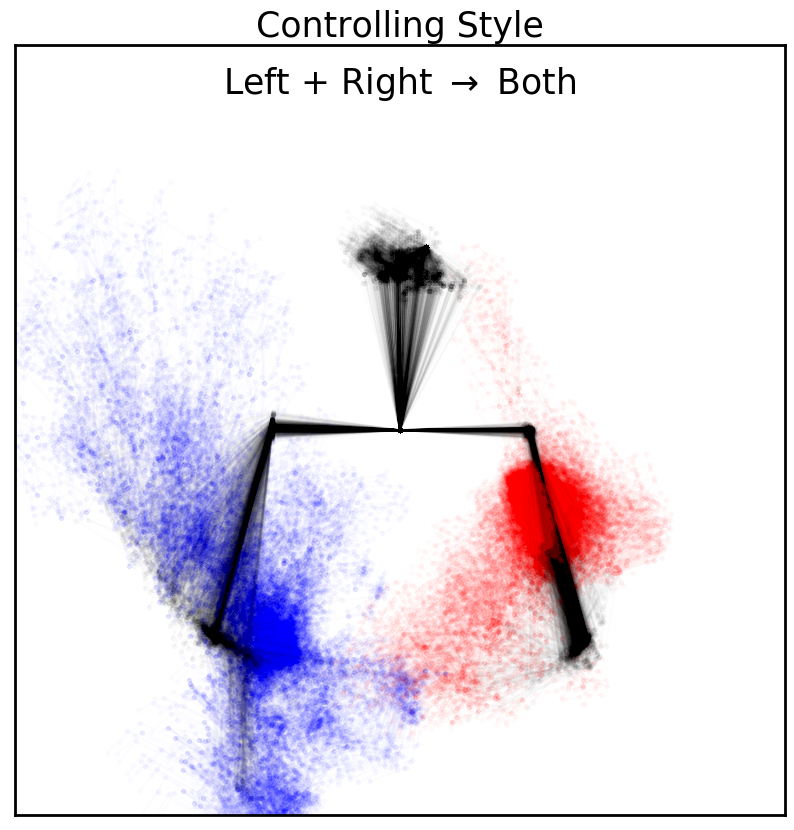}
\end{center}
\caption{Heat map of mixture of two styles: primary arm function of left and right mixed to give motion for both hands. Red represents the left hand and blue represents the right hand.}
\label{fig:mix}
%\end{figure}
\end{figure}

As a preliminary experiment, we modified the style vector to $[0.5, 0.5]$ in order to mix the styles of two speakers with different Primary Arm Functions. The generated gesture space in Figure \ref{fig:mix} indicates that different speaker styles could be interpolated into a completely new style. 

\subsubsection{Future Directions}
Our efforts were aimed at modeling, disentangling and transferring gesture style under the assumption that the emotional state of the speaker does not affect the gestures. While this is reasonable for speakers in \datasetshortss, which are mostly scripted monologues, it may not be true in general hence motivating an interesting future direction. Another direction, that might induce diversity in the generated gestures, is the inclusion of verbal information (i.e. natural language). This may not be trivial in context of style transfer as the difference in vocabulary of different speakers could create an unwanted bias - some words might get associated with certain styles of gesturing.

% \subsection{Prototypical gestures}
% Prototypical gestures of different speaker visualized as heatmaps in Figure \ref{fig:clusters_all}. Each of the prototypical gesture represents one of the \textit{modes} in the underlying probability distribution of poses. Each row corresponds to one speaker and with different \textit{modes} every column. It is important to note that \textit{mode} $m$ of two speakers can represent completely different distributions, as these partitions were estimated independently in an unsupervised manner. These figures are useful for determining the diversity in gestures for a given speaker.
% \input{sup_z99_clusters.tex}
% %\input{sup_z94_moca_m.tex}

\section{Implementation Details}
This section gives more detail about the exact architectures used for our model also described in Section 4 of the main paper.
\subsection{Network Architectures}
\begin{figure}[h]
\includegraphics[width=\linewidth]{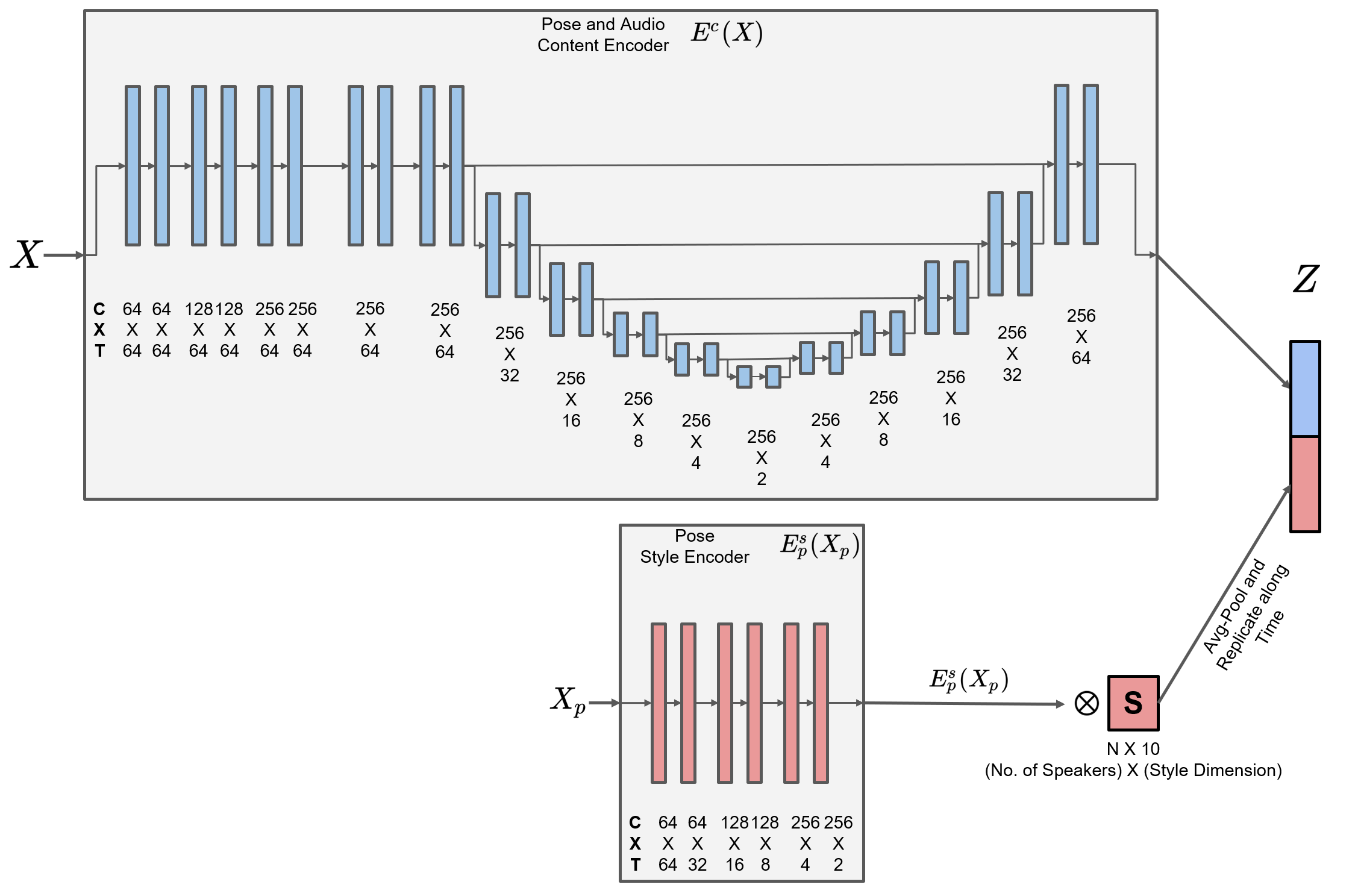}
\caption{Encoder Architecture} \label{fig:encoder}
\end{figure}

\begin{figure}[h]
\centering
\includegraphics[width=0.5\linewidth]{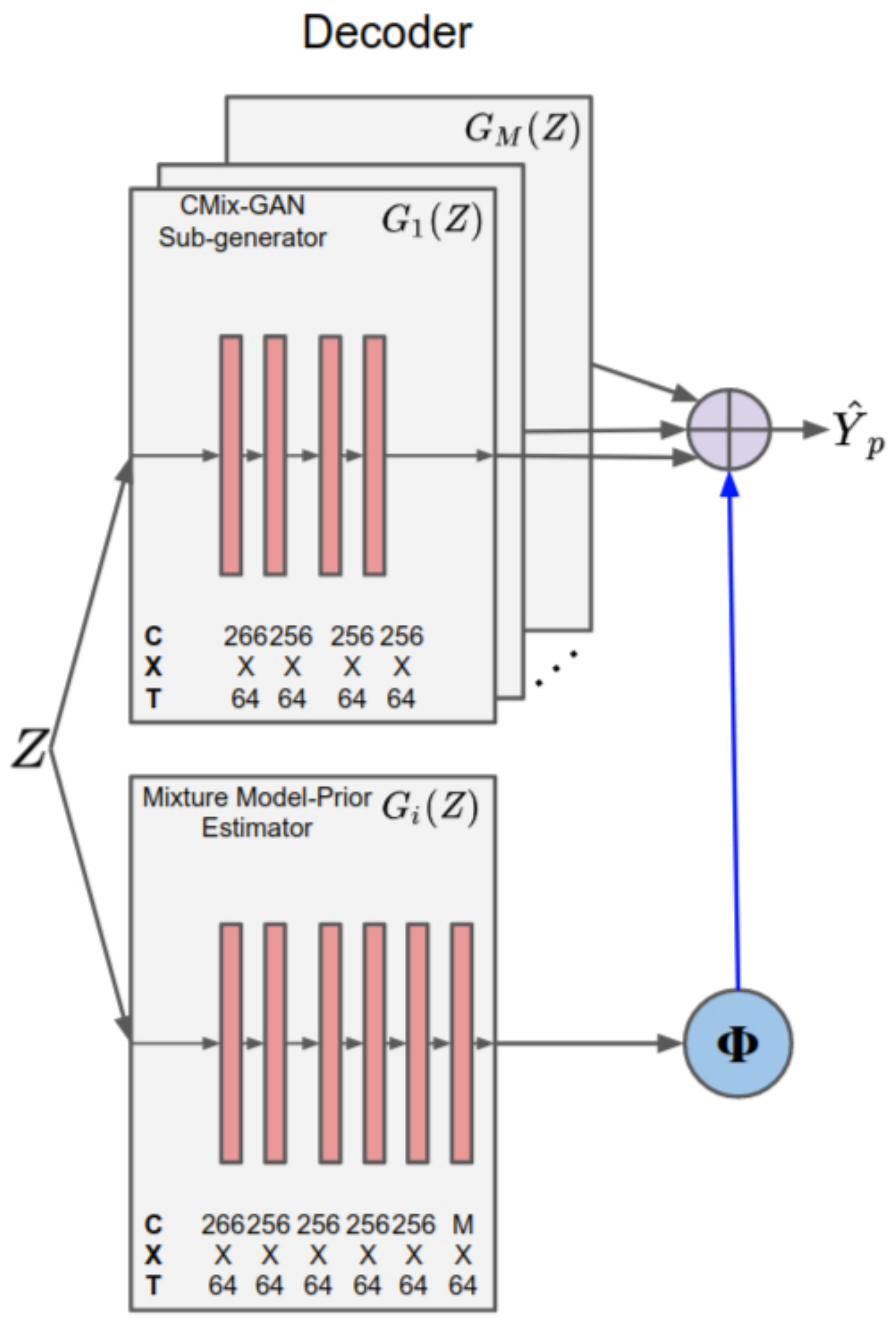}
\caption{Decoder Architecture} \label{fig:decoder}
\end{figure}

Figure \ref{fig:encoder} and \ref{fig:decoder} consists of the visual representation of the architectures used for our model \modelshorts. For the decoder $\bigoplus$ is a weighted sum as described in Equation (2) of the main paper. Every operation is a \textbf{1D-convolution} followed by a \textbf{Batch-Norm} and finally \textbf{ReLU}. Each convolution uses a kernal size of 3 and hop length of 1, except for cases where temporal dimension is downsampled where the kernal size is 4 and hop length is 2.

\subsection{Training Details}
We use Adam \cite{kingma2014adam} to optimize the model with a exponentially decaying learning rate of $0.001$. We train each model for 60000 iterations while check-pointing every 3000 iterations. Finally, we choose the best model based on loss on the development set. We use $\lambda_{id} = 0.1$ to prevent the style consistency loss from stealing focus while training the pose gesture generator.

\subsection{Human study experiments}
We conducted our human studies on Amazon Mechanical Turk (or AMT), for which we used 100 random videos for each speaker which gave us 2400 pairs of comparisons per model for each study (out of 5). This is a significant number of comparisons and helps with reliability of the results. Each annotation task contained 20 videos and is performed by 3 different users; hence we had approximately (2400*3/20) = 360 participants.

To help filter unreliable annotators, we use two ground truth videos from the same speaker with the same style as control samples. If annotators tag these two videos as different styles, then we disregard this annotation set as unreliable.\\

\noindent \textbf{Sample study for style transfer (other studies also follow similar method)}\\
Two videos are shown to the user. One video is a ground truth(Speaker A Style A) and the other is generated by a model. The generated video could be either of (a) Speaker A Style A or (b) Speaker B Style A.
We ask two questions to measure correctness of style transfer and naturalness: 
\begin{enumerate}
    \item Do the animations have different styles of gestures?
    \item Which of the videos 1 or 2 has more Natural gestures with respect to the audio?
\end{enumerate}

\section{Videos: Style Transfer and Preservation}
We refer the readers to \url{http://chahuja.com/mix-stage} for demo videos.
%These results are videos and complement Figure 7 in the main paper. We urge the readers to check out the file \textbf{appendix\_b.html} and \textbf{appendix\_c.mp4} in the supplementary packet, which contains videos demonstrating Style Transfer and Style Preservation animations respectively. 

\section{Video Frames: Style Preservation Qualitative Results}
\begin{figure}[!h]
\centering
\caption{Animation depicted as a series of frames for different speakers. The vertical axis is labeled as models and horizontal axis is time. The generated animation is superimposed over the ground truth video.} \label{fig:animations}
\begin{subfigure}[t]{\textwidth} \centering
\includegraphics[width=\linewidth]{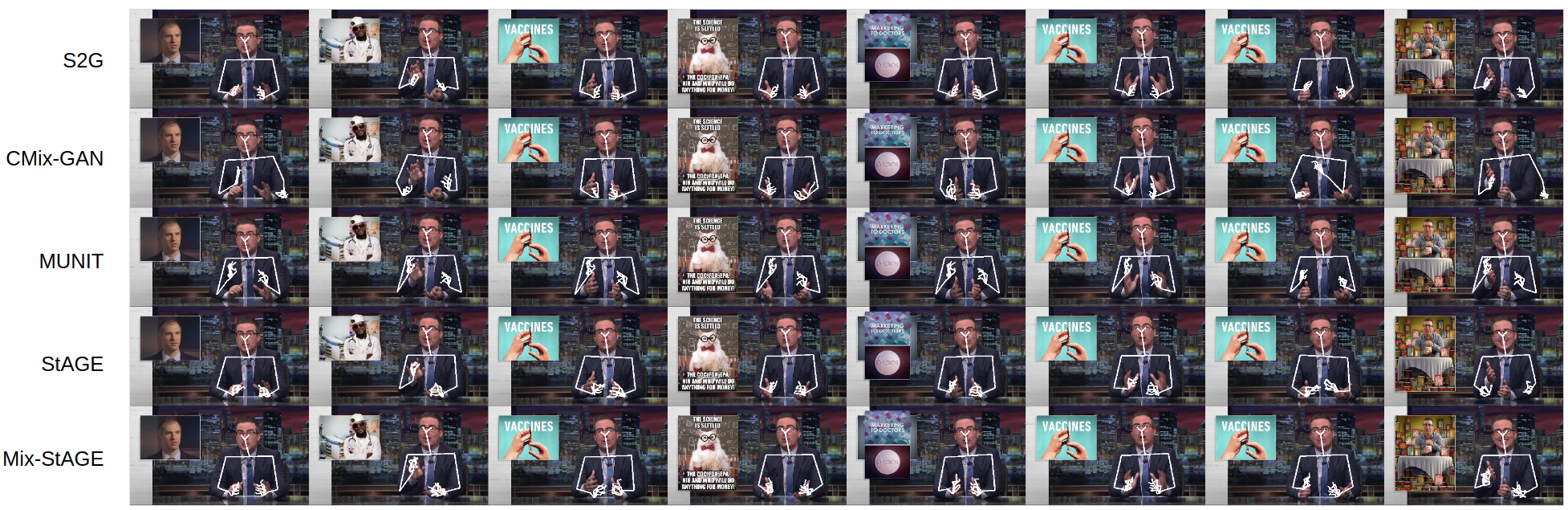}
\vspace{-0.5cm}
\caption*{oliver}
\vspace{0.1cm}
\end{subfigure}
\begin{subfigure}[t]{\textwidth} \centering
\includegraphics[width=\linewidth]{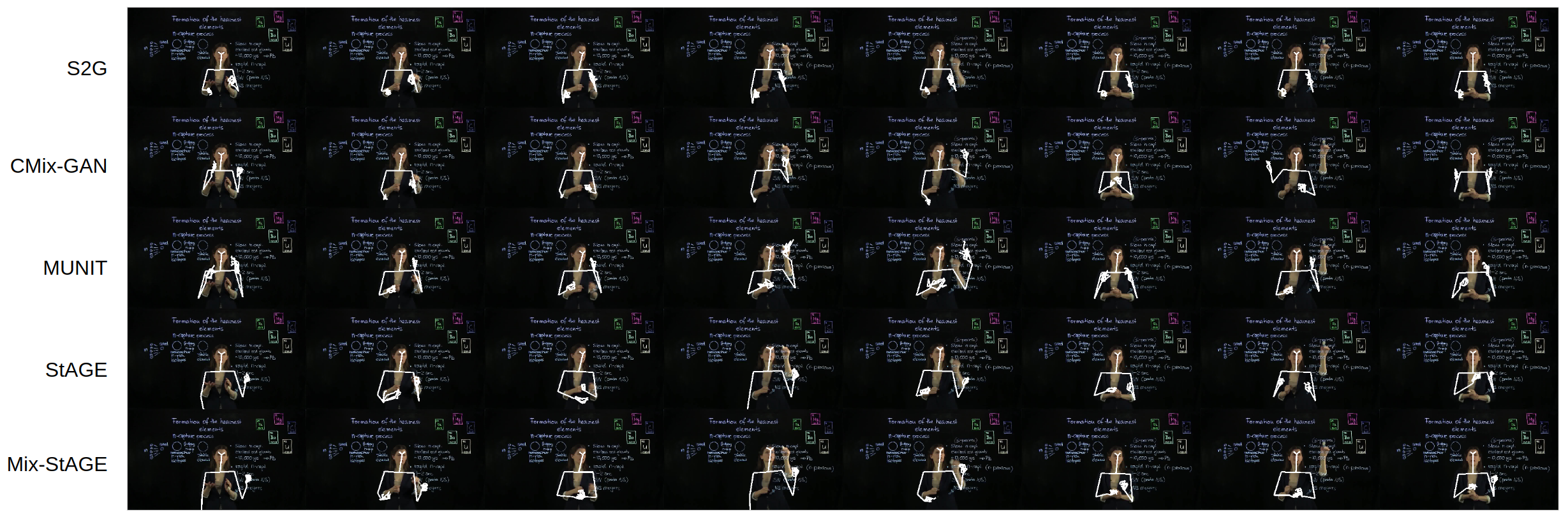}
\vspace{-0.5cm}
\caption*{lec\_cosmic}
\vspace{0.1cm}
\end{subfigure}
\begin{subfigure}[t]{\textwidth} \centering
\includegraphics[width=\linewidth]{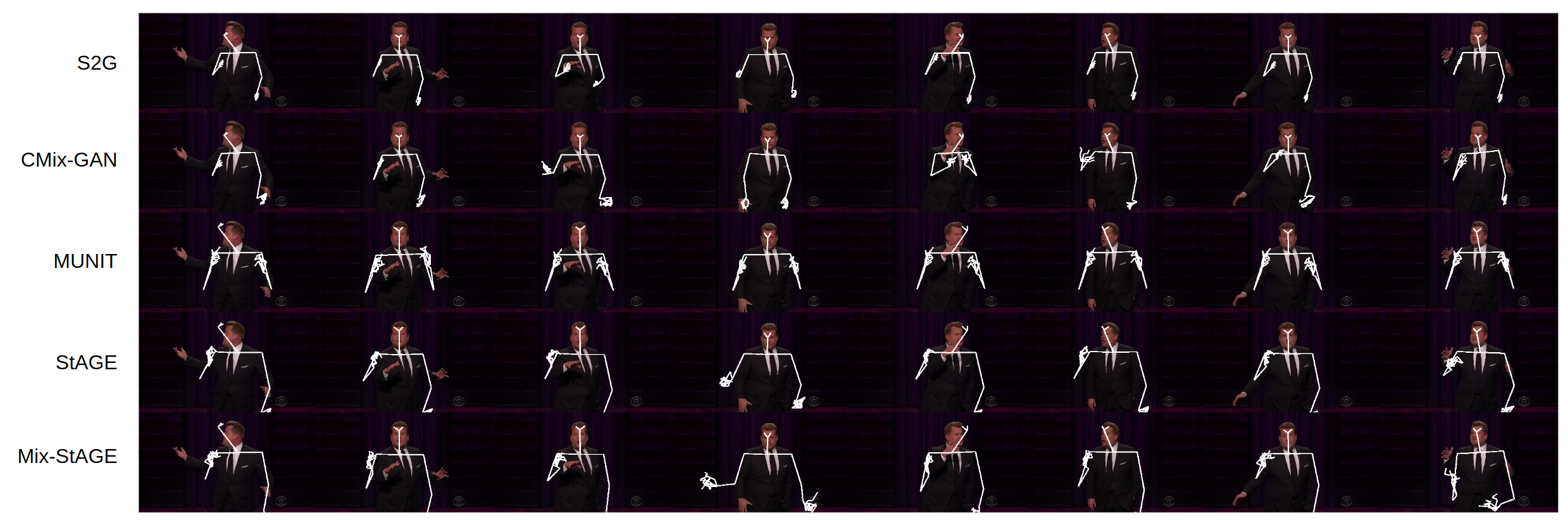}
\vspace{-0.5cm}
\caption*{corden}
\vspace{0.1cm}
\end{subfigure}
\begin{subfigure}[t]{\textwidth} \centering
\includegraphics[width=\linewidth]{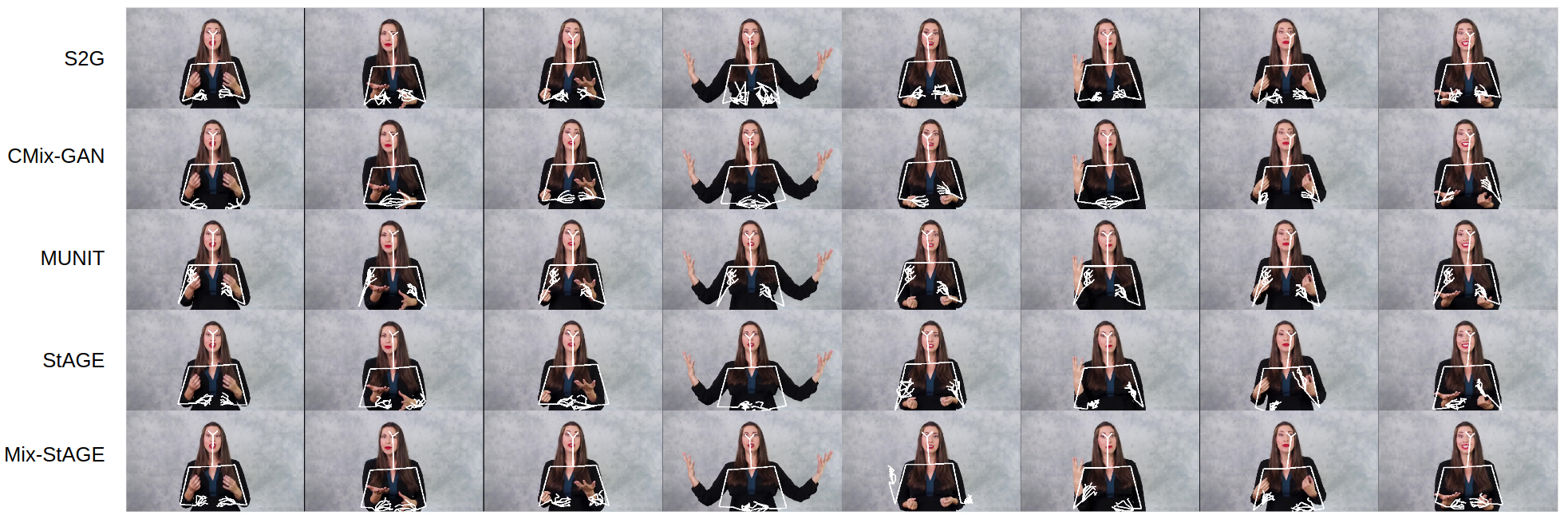}
\vspace{-0.5cm}
\caption*{ytch\_prof}
\vspace{0.1cm}
\end{subfigure}
\end{figure}

These results complement Figure 8 in the main paper. We plot some more animation figures generated by random audio samples in the test-set to provide some more samples for qualitative judgment in Figure \ref{fig:animations}.

\end{document}